\documentclass[journal]{IEEEtran}
\ifCLASSINFOpdf
\else
\fi

\usepackage{booktabs}
\usepackage{graphicx}
\usepackage{subcaption}
\usepackage[acronym]{glossaries}
\usepackage{amsmath,amsfonts}
\newacronym{radar}{RADAR}{RAdio Detection And Ranging}
\newacronym{lidar}{LIDAR}{LIght Detection and Ranging}
\newacronym{lstm}{LSTM}{Long Short-Term Memory}
\newacronym{cnn}{CNN}{Convolutional Neural Network}
\newacronym{knn}{K-NN}{K-Nearest Neighbor}
\newacronym{mlp}{MLP}{Multi-Layer Perceptron}
\newacronym{tfn}{TFN}{Transformer Network}
\newacronym{mpnn}{MPNN}{Message Passing Neural Network}
\newacronym{svm}{SVM}{Support Vector Machine}
\newacronym{gflop}{GFLOP}{Giga Floating Point OPeration}
\newacronym{mb}{MB}{Mega Bytes}
\newacronym{ahc}{AHC}{Agglomerative Hierarchical Clustering}
\newacronym{iot}{IoT}{Internet of Things}
\newacronym{tfnet}{TFNet}{Transformer Network}
\newacronym{gpu}{GPU}{Graphical Processing Unit}
\newacronym{adc}{ADC}{Analog to Digital Conversion}
\newacronym{gnn}{GNN}{Graph Neural Network}
\newacronym{nlp}{NLP}{Natural Language Processing}
\newacronym{rnn}{RNN}{Recurrent Neural Network}
\newacronym{fmcw}{FMCW}{Frequency-Modulated Continuous Wave}
\newacronym{mimo}{MIMO}{Multiple-Input and Multiple-Output}
\newacronym{fft}{FFT}{Fast Fourier Transform}
\newacronym{cfar}{CFAR}{Constant False Alarm Rate}
\newacronym{ubpg}{UBPG}{Upper Body Point Cloud Gestures}
\newacronym{rf}{RF}{Radio Frequency}
\newacronym{dec}{DEC}{Dynamic Edge Convolution}
\newacronym{auc}{AUC}{Area Under ROC Curve}
\newacronym{rl}{RL}{Reinforcement Learning}
\newacronym{csi}{CSI}{Channel State Information}
\begin{document}
%
\title{Integrating Sensing and Communication in Cellular Networks via NR Sidelink}
%
%
%

\author{Dariush~Salami,~\IEEEmembership{Member,~IEEE,}
        Ramin~Hasibi, Stefano~Savazzi~\IEEEmembership{Member,~IEEE,}
            Tom~Michoel, and~Stephan~Sigg,~\IEEEmembership{Senior~Member,~IEEE}
\thanks{D. Salami and S. Sigg are with the Department of Communications and Networking, Aalto University, Finland.\protect\\
E-mails: \{dariush.salami, stephan.sigg\}@aalto.fi}
\thanks{R. Hasibi and T. Michoel are with the Department of Informatics at the University of Bergen.\protect\\
E-mails: \{ramin.hasibi, tom.michoel\}@uib.no}
\thanks{S. Savazzi is a with the National Research Council of Italy (CNR), IEIIT institute, Milano, Italy.\protect\\
E-mail: stefano.savazzi@ieiit.cnr.it}
}

%
%

\markboth{IEEE Journal on Selected Areas in Communications,~Vol.~XX, No.~XX, August~2021}%
{Shell \MakeLowercase{\textit{et al.}}: Bare Demo of IEEEtran.cls for IEEE Journals}
%



\maketitle

\begin{abstract}
RF-sensing, the analysis and interpretation of movement or environment-induced patterns in received electromagnetic signals, has been actively investigated for more than a decade. 
Since electromagnetic signals, through cellular communication systems, are omnipresent, RF sensing has the potential to become a universal sensing mechanism with applications in smart home, retail, localization, gesture recognition, intrusion detection, etc. 
Specifically, existing cellular network installations might be dual-used for both communication and sensing. 
Such communications and sensing convergence is envisioned for future communication networks. 
We propose the use of NR-sidelink direct device-to-device communication to achieve device-initiated, flexible sensing capabilities in beyond 5G cellular communication systems. 
In this article, we specifically investigate a common issue related to sidelink-based RF-sensing, which is its angle and rotation dependence. 
In particular, we discuss transformations of mmWave point-cloud data which achieve rotational invariance, as well as distributed processing based on such rotational invariant inputs, at angle and distance diverse devices.
To process the distributed data, we propose a graph based encoder to capture spatio-temporal features of the data and propose four approaches for multi-angle learning. 
The approaches are compared on a newly recorded and openly available dataset comprising 15 subjects, performing 21 gestures which are recorded from 8~angles.
\end{abstract}


%
\IEEEpeerreviewmaketitle

\section{Introduction}
%
%
%
%
\IEEEPARstart{R}{F} convergence describes the use of a common transceiver for both communication and sensing. 
In particular, the recent advances in RF-sensing, such as localization and recognition has demonstrated that ubiquitous cellular installations may not only enable seamless connectivity, but, utilizing the same system, it may further seamless localization and environmental perception capabilities. 
However, as the available bandwidth is naturally limited~\cite{rohde2021detection}, RF-sensing has to share resources with communication systems. 
In particular, as shown in~\cite{herschfelt2020vehicular}, resource sharing is more efficient than reserving fixed-size resources for both communication and sensing. 
In this regard, modulating communication symbols with radar waveforms, as well as the utilization of communication signals for localization and environmental perception have been investigated~\cite{bicua2019multicarrier}. 

\begin{figure}
 \includegraphics[width=\columnwidth]{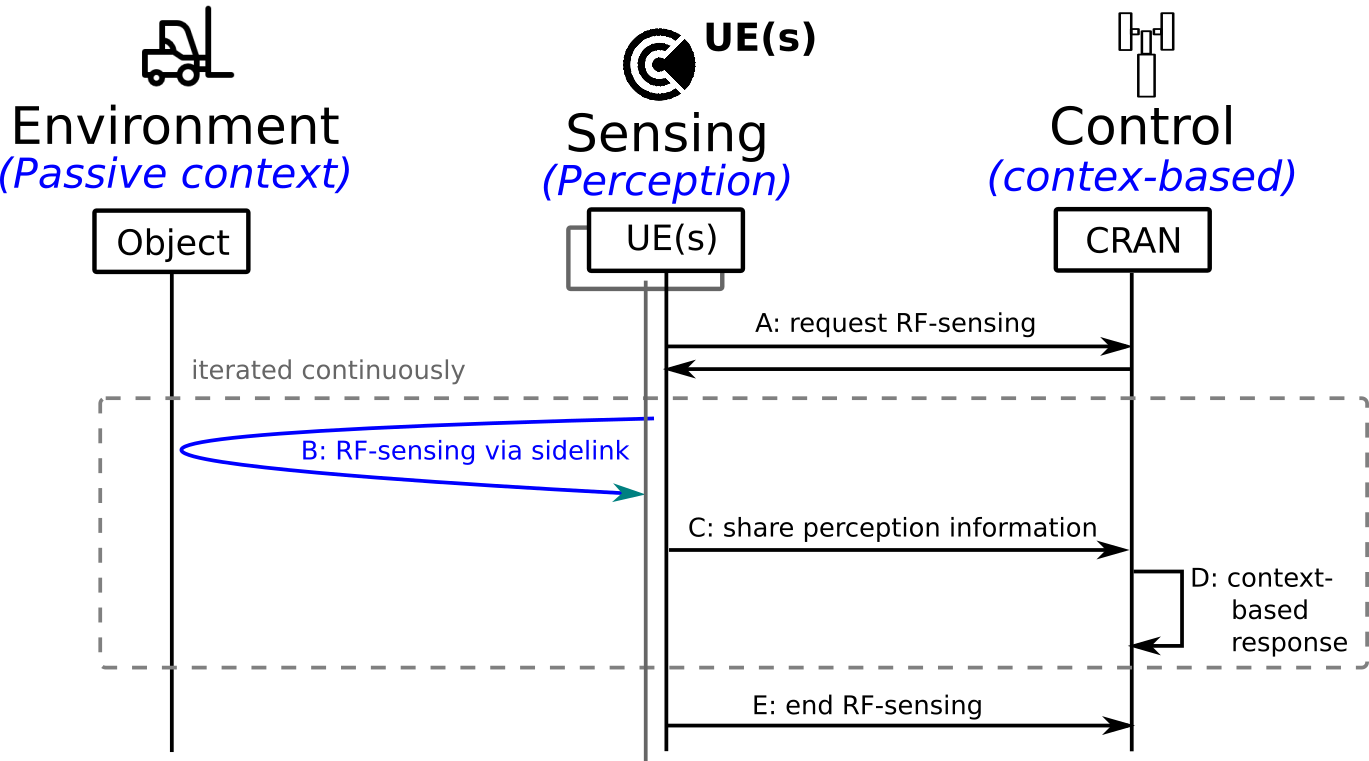}
  \caption{Integration of sensing capabilities into next generation cellular communication systems exploiting sidelink device-to-device communication}
  \label{fig:teaser}
\end{figure}

In view of the current standardization efforts, neither of these approaches were deeply investigated for future communication systems. 
However, 3GPP has introduced a mechanism for devices to reserve channel resources in the system: the LTE/NR sidelink feature (PC5 interface), which has been introduced first in LTE releases 12 and 13 for device-to-device communication. 
Specifically, a user equipment (UE) device may request resources for sidelink communication from the corresponding gNB. 
According to the standard, the sidelink communication then occurs via the Physical Sidelink Shared Channel (PSSCH). 
We propose to utilize the sidelink functionality for device-based RF-sensing (cf. figure~\ref{fig:teaser}). 
In particular, a UE would request sidelink resources from the respective gNB, and then operates, for instance, an FMCW radar in the assigned frequency range\footnote{Both licensed and unlicensed bands may be allocated for NR sidelink use}.

In addition, we investigate and address a practical issue in RF-sensing systems, specifically in distributed sidelink-sensing settings, which is frequently ignored in the literature. 
The relative orientation of the RF-interface and the subject (e.g. in gesture sensing) or (moving) object (e.g. environmental perception) significantly impacts the observed electromagnetic pattern and hence the recognition accuracy. 
We demonstrate this impact in an experimental study with 15~subjects, in which gestures are simultaneously recorded from 8~different angles. 
A learning system is proposed which is capable of recognizing gestures from multiple angles. 
For this, we employ in a sidelink-sensing scenario multi-angle learning, four different pooling mechanisms to weigh the relevant angles, as well as rotationally independent representations of the received patterns. 
The system is compared to traditional RF-sensing approaches and shows superior recognition accuracy. 

Our contributions are
\begin{itemize}
 \item a scheme to achieve RF-convergence in cellular communication systems utilizing NR sidelink for flexible UE-initiated and CRAN controlled resource allocation.
 \item four approaches for multi-angle learning to fuse data from distributed sidelink-based RF-sensors
 \item a novel graph based encoder to capture spatio-temporal features of point clouds
 \item an openly available dataset for a NR sidelink scenarios collected from 15 subjects from 8 angles with more than 25,000 samples
 \item an open source data collection platform\footnote{https://version.aalto.fi/gitlab/salamid1/mmwave-data-collector} for mmWave radars as well as the source code of the proposed models \footnote{https://version.aalto.fi/gitlab/salamid1/AngleRadar} 
\end{itemize}

The proposed system for RF-convergence via NR sidelink will enable simultaneous communication and sensing and seamlessly integrates into future cellular communication networks by widening the sidelink communication capabilities.   

Possible applications, for instance, may be the support of ultra-reliable and low-latency communication (URLLC) through pro-active corrective actions (e.g. switching the serving gNB ahead of a predicted signal drop).
Particularly, the system may exploit RF sensing for channel prediction~\cite{chen2019contactless,chen2020robust,savazzi2019use}. 
RF-data provides rich features to predict sudden events (e.g., blockages), which cannot be done with current, purely channel analytic solutions.
Further examples are ambient assisted living scenarios, gaming, intrusion detection, as well as elderly/remote care, where pre-installed cellular communication hardware may be utilized for the recognition of situations, gestures, motion and mobility.

\section{Related Work}
Device-free gesture and activity recognition, by analysing time or frequency domain patterns of electromagnetic signals, achieves high recognition accuracy~\cite{savazzi2016device}. 
For instance, the Channel State Information (CSI), phase and Received Signal Strength (RSS) on channels of multi-antenna wireless interfaces provide rich information and enable accurate recognition of motion~\cite{ma2019practical,palipana2021pantomime}. 
In particular, via micro Doppler variations, whole-body motion can be distinguished~\cite{skaria2019hand}. 
Furthermore, detection of minute movement, such as respiration, pulse~\cite{wang2020resilient} or even emotion~\cite{raja2018towards} may be obtained using Frequency Modulated Continuous Wave (FMCW) radars. 
In addition, via RF-holography, the 3D perception of objects~\cite{holl2017holography} is feasible. 
The technology is capable of joint motion detection for multiple moving targets~\cite{zhao2018through} by signal processing to separate motion of these targets from a single stream and by installing multiple Tx/Rx points.

These impressive results, together with the realization that electromagnetic signals are omnipresent nowadays through cellular system deployment, explain the interest in integrating sensing capabilities into communication systems. 
A good overview on RF communications and sensing convergence research is given in~\cite{paul2016survey}. 
In particular, the authors argue that the wireless medium is limited and thus has to be shared for both communication and sensing systems. 
For instance, DARPA has proposed shared spectrum access for radar and communications\footnote{http://www.darpa.mil/program/shared-spectrum-access-for-radar-and-communications}. 
They propose cooperative spectrum sharing, in which information is shared between the communication and the radar system in near real time, which implies the co-existence of two separate systems.  
In contrast, to achieve communication and sensing via a single, integrated system, several authors have proposed communication that is embedded in signals transmitted by radar devices (radar-embedded communications)~\cite{blunt2010intrapulse,parmentar2019opportunistic,metcalf2020impact,liu2021deep}. 
For instance,~\cite{bicua2019multicarrier} propose a dual-use radar-communication multicarrier waveform, in which different subcarriers are assigned to different subsystems to achieve RF-convergence. 
In contrast, the authors of~\cite{rohde2021detection} investigate the use of guard bands of a linear frequency modulated radar waveform for communications and study how the waveform affects the symbol error rate.  
Additionally, they also derive the radar's probability of detection when the QPSK RF carriers are injected into the radar signal. 
A good overview over recent results on RF convergence is given in~\cite{herschfelt2020vehicular}, together with a proposal for a joint radar, communications, positioning, navigation and timing system.


In contrast to this previous work, we suggest to utilize the NR sidelink device-to-device communication functionality that has been first introduced by 3GPP in releases 12 and 13 for LTE. 
In particular, we suggest that UEs request resources from the system for RF sensing to instrument mmWave radar sensing on the granted resources. 
Hence, spectrum sharing would be dynamically adaptive to the RF-sensing need, indicated autonomously by UE devices and controlled by the system to balance communication and sensing needs in any given situation. 

Recently, a related concept has been discussed in~\cite{lu2021joint}. 
The authors propose to continuously track neighbouring devices by estimating angle of arrival (AoA), time of arrival (ToA) and received signal strength (RSS) from NR sidelink reference signals exchanged between the devices. 
In contrast, we propose to use the bandwidth granted for NR sidelink communication for mmWave radar sensing (e.g. FMCW), instead of utilizing the packet-based communication through the shared and controlled sidelink channels PSCCH and PSSCH. 


In our work we conduct an RF-sensing instrumentation utilizing the Texas Instruments IWR1443\footnote{https://www.ti.com/product/IWR1443} sensor, which is a \gls{fmcw}–\gls{mimo} radar operating in the 77-81 GHz RF band. 
In particular, the sensor generates point clouds reflected from objects in the environment to yield a time-varying signal in an $x$-$y$-$z$ coordinate system~\cite{palipana2021pantomime}.

A temporal point-cloud is a sequence of frames through time, each of which consists of an unordered set of points~\cite{min2020efficient, salami2020motion}. 
Fuelled by commercial availability of point cloud generating mmWave radars, mmWave radar point cloud based sensing has been actively investigated for e.g. hand tracking~\cite{dong2020model}, gesture recognition~\cite{liu2020real} activity recognition~\cite{singh2019radhar}, gait recognition~\cite{meng2020gait}, or positioning~\cite{zhao2019mid}.

Direct point cloud processing has first been achieved by PointNet~\cite{qi2017pointnet}, which extracts point-wise spatial features and aggregates them through permutation-invariant pooling operations. 
To tackle the problem of learning local structures in PointNet, PointNet++~\cite{qi2017pointnet2} was proposed. 
In particular, the authors have introduced a Set Abstraction (SA) layer to recursively apply a simplified version of PointNet on the input point cloud capturing both local and global features. 
Following a graph based approach,~\cite{wang2019dynamic} proposed Dynamic Edge Convolution (DEC) to build graphs by applying a nearest neighbour search on point clouds and by processing them using Message Passing Neural Networks (MPNN). 
Although these models have shown significant improvements on static point cloud processing including shape classification and semantic segmentation, they lack the ability to capture temporal dependencies such as in gestures and movement.

To process dynamic point clouds, a combination of Recurrent Neural Networks (RNNs) with either 3D-Convolutional Neural Networks (CNNs) or PointNet++ layers has been proposed in~\cite{palipana2021pantomime,salami2020motion,owoyemi2018spatiotemporal}. 
Furthermore, a modified LSTM layer,~\cite{min2020efficient}, was introduced to propagate temporal information while preserving the spatial structure.
All these models, however, expect a point cloud from a single sensor and are trained with respect to a single unique operating angle with respect to the observed gesture or movement. 
Naturally, single angle and single sensor operations are constrained and constitute only a special case in realistic instrumentation, where multiple sensors in an environment might contribute point clouds of a scene from various different angles. 
In such case, synchronized operation across sensors is necessary and angles might deviate from the optimal angle used for training. 
We propose models which are resilient to changes in angle and which are also capable to operate on varying number of input angles for environmental perception.


\section{RF-sensing for cellular communications}
RF convergence describes the shared use of the available bandwidth in a communication system for both communication and sensing. 
Traditional approaches have designed signals to either embed communication in transmissions by radar devices, or have utilized communication signals for sensing~\cite{herschfelt2020vehicular}. 
In both cases, a degradation in performance compared to a communication-only (sensing-only) system is expected~\cite{rohde2021detection}.

However, the demand in sensing or communication services differs in its distribution both temporally and spatially. 
Especially in cellular systems, the RF-sensing demand which is short-range, is primarily associated to UEs while a gNB has a higher communication demand since it connects to multiple UEs. 
For efficiency reasons, resources reserved for UE-centered spatially constrained sensing should not constrain communication or RF-sensing of remote UEs in the same cell. 
A mechanism that supports this paradigm is the NR-sidelink device-to-device communication. 
In particular, in NR-sidelink, a UE may request resources for spatially constrained sidelink operation. 
We propose to implement RF-sensing utilizing FMCW on the reserved sidelink resources. 

In the following, we briefly introduce the specifics of NR sidelink communication, further propose RF-sensing via sidelink, and discuss angle dependency in NR sidelink based RF-sensing.

\begin{figure*}
 \includegraphics[width=\textwidth]{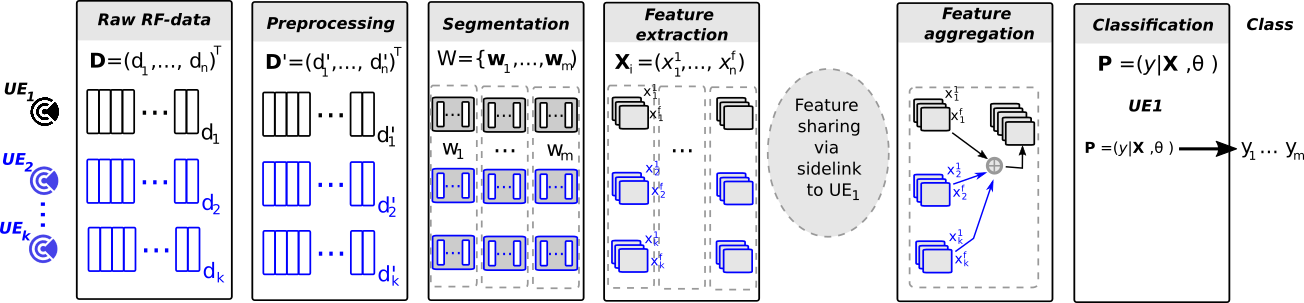}
 \caption{Data processing and aggregation chain for multiple devices in proximity conducing joint NR-sidelink-based RF sensing}
 \label{fig:processingChain}
\end{figure*}
\subsection{NR sidelink (R17)}
3GPP introduced the LTE sidelink feature (PC5 interface) in release 12 and 13 for device to device communication with minor involvement of the eNB~\cite{molina2018configuration} (device-to-device (D2D) proximity service (ProSe)). 
In particular, two or more UEs in proximity can be known by each other using direct signalling and reserved resources. 
To enable the functionality, a new node, ProSe (Proximity Services) function is introduced of which each PLMN possesses one~\cite{lin2014overview}. 
Authorization to use direct discovery or direct communication of a UE is done by this node. 
The ProSe function provides Radio resource parameters to the UEs for out of coverage communication. 
UEs are further capable of direct discovery (identifying another UE in proximity) via E-UTRA direct radio signals.
Sidelink communication is conducted via the Physical Sidelink Shared Channel PSSCH (for ProSe application data; same as LTE PDSCH, QPSK \& 16QAM), the Physical Sidelink Control Channel PSCCH (for control information to decode PSSCH, similar to PDCCH, QPSK), Physical Sidelink Discovery Channel PSDCH and Physical Sidelink Broadcast Chanel PSBCH (for MIB-SL information). 
Resource pools are set of resources that are used for sidelink transmission. 
Both control and data information is transmitted using a resource pool. 
The Radio resources for direct communication can be requested by the UE autonomously or given by the gNB. 
To initialize device-to-device communication, the UE sends ProSe UE information indication informing the network that it wants to use direct communication. 
gNB will in turn give radio resource information in an RRC message along with SL-RNTI for that UE to decode DCI\footnote{\label{3GPPTR23725}3GPP TR 23.725 V16.2.0 (2019-06), 3rd Generation Partnership Project; Technical Specification Group Services and System Aspects; Study on enhancement of Ultra-Reliable Low-Latency Communication (URLLC) support in the 5G Core network (5GC)(Release 16)}. 
The gNB then assigns the dedicated resources to the UE and sends a grant using DCI\footnotemark[\value{footnote}] scrambled with SL-RNTI and the UE uses this information to send data in sidelink to UE2 using SCI 0 for control information before data is sent via PSSCH between a pair of UEs.

3GPP introduced in their Release 15 of the 5G new radio (NR) specification the Sidelink technology for cellular vehicle-to-everything communication (C-V2X). 
In the following release 16, physical layer, protocol and core network functionalities supporting NR sidelink for Broadcast, Groupcast, and Unicast were added together with positional reference signal (PRS) and a sounding reference signal (SRS)~\footnote{\label{3GPPTR22186}3GPP TR 22.186 V16.2.0 (2019-06), 3rd Generation Partnership Project; Technical Specification Group Services and System Aspects; Enhancement of 3GPP support for V2X scenarios; stage 1 (Release 16)}. 
The 3GPP Release 17 standardization is investigating enhancements to NR sidelink to support higher reliability for V2V services, power saving features for pedestrian/vulnerable Road Users, commercial D2D use-cases etc. 
The NR Release 16 sidelink design assumes two frequency ranges, up-to 6 GHz and a Subcarrier Spacing (SCS) of 15~KHz, 30 KHz and 60~KHz, as well as a frequency range from 6 GHz to 52.6~GHz and with a SCS of 60~KHz and 120~KHz. 

The Physical Sidelink Control Channel (PSCCH) carries Sidelink Control Information (SCI) and is time multiplexed with the associated Physical Sidelink Shared Channel (PSSCH). 
Furthermore, a two-stage SCI was introduced, separating the content of the SCI payload into two parts. 
The first SCI part contains sensing information and is broadcast to the surrounding UEs (containing information related to QoS priority of data, occupied resource blocks, resource reservation interval etc.), whereas the second SCI transmission carries information related data decoding of Physical SL Shared Channel (PSSCH)~\cite{wang2017overview}. 
Furthermore, higher reliability for unicast and groupcast transmission is achieved with the introduction of the new Physical Sidelink Feedback Channel (PSFCH) to carry HARQ feedback in the last OFDM symbol of a slot. 
In order to perform efficient link adaptation for unicast transmissions, feedback of the channel quality information (CQI) is supported from the receiver UE.

\subsection{NR sidelink-based RF-sensing}
\begin{figure}
    \centering
    \includegraphics[width=\columnwidth]{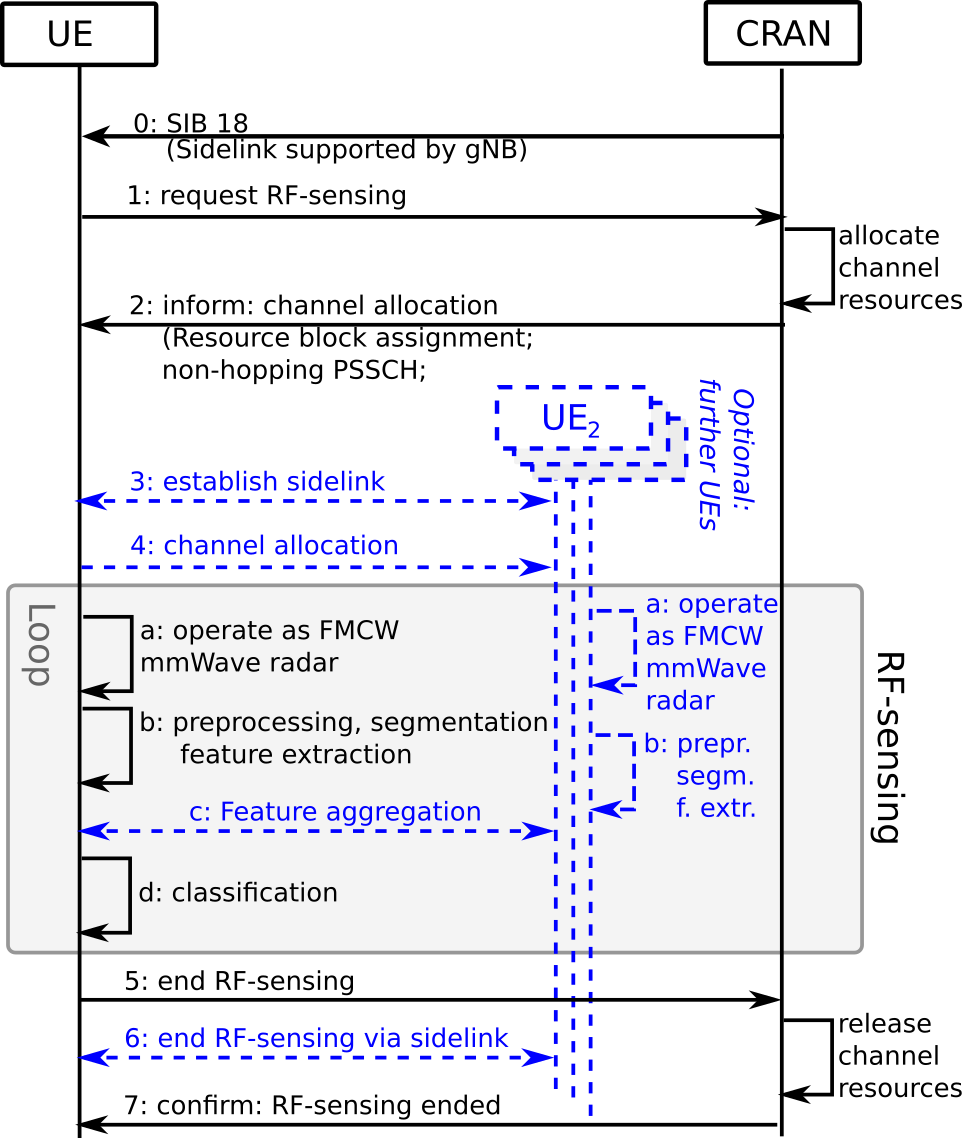}
    \caption{RF-convergence achieved via the allocation of sidelink resources and FMCW mmWave sensing in a cellular communication system. In the case of multiple participating UEs for the sensing (indicated in blue in the figure), feature aggregation and communication across UEs is achieved via a sidelink communication channel. }
    \label{fig:sidelinkSensing}
\end{figure}

NR-sidelink is designed to support D2D communication utilizing a battery of the usual broadcast, control, shared and feedback channels (PSBCH, PSCCH, PSSCH) together with a number of reference signals (primary (S-PSS), secondary (S-SSS), phase tracking (PF-RS), and CSI-RS). 
NR sidelink is expected to work in both licensed and unlicensed bands and operates out-of-coverage, in partial coverage or in-coverage. 
We propose a new mode of operation for NR-sidelink: environmental sensing. 
In particular, a UE would request resources for RF-sensing from the gNB or CRAN, which in turn informs the UE about the assigned resource block (steps 1 and 2 in figure~\ref{fig:sidelinkSensing}). 
The UE may then utilize these resources, e.g. operating as a FMCW mmWave radar to sense an environment and gestures (steps a-d in the figure).
If multiple UEs in proximity participate in the sensing of an environmental situation, the initiating UE first establishes a sidelink communication channel among the UEs, and shares information on the allocated channel resources (steps 3 and 4). 
In such case, the features extracted are to be aggregated across these devices (step c). 
After the sensing operation is completed, the UE informs the gNB that the resources may be released again (cf. steps 5-7 in figure~\ref{fig:sidelinkSensing}).
The environmental sensing processing chain for each UE participating in the sidelink sensing is illustrated in figure~\ref{fig:processingChain}. 
In the case of a single UE, the device extracts the data, processes and segments it and utilize the extracted features for the classification.
When multiple UEs collaborate (in blue in the figure) by jointly sensing an environmental situation, each device will extract the data, preprocess, segment it and extract features from the local data stream. 
These features are then shared via the sidelink with the initiating UE, which in turn aggregates the features and infers the respective class (cf. figure~\ref{fig:processingChain}). 

\subsection{Angle dependency in RF-sensing}
\begin{figure}
 \includegraphics[width=\columnwidth]{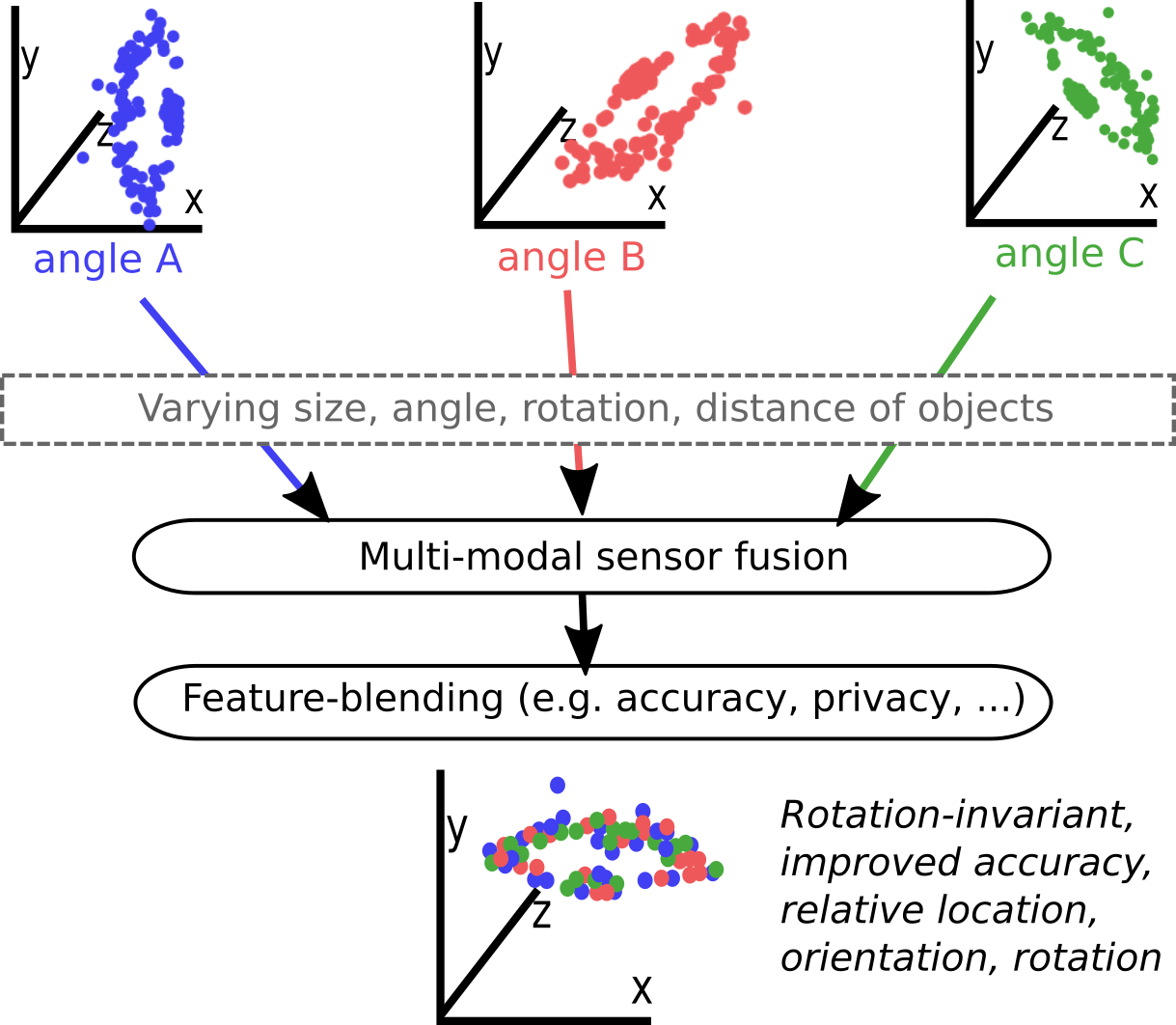}
 \caption{Illustration of the fusion of multi-angle point-cloud data. Clapping gestures recorded with the mmWave radar simultaneously from different angles are merged into a join representation of the gesture.}
 \label{fig:multi-sensorFusion}
\end{figure}

RF-sensing collaboration between multiple devices in proximity touches a fundamental problem in RF-sensing. 
The observed patterns which originate from reflections off the same moving object differ among sensing UEs conditioned on their distance, angle and translation relative to the object. 
A gesture recorded from a $0^\circ$ angle might have a very different shape when recorded e.g. from a $90^\circ$ angle as illustrated in figure~\ref{fig:multi-sensorFusion}. 
In order to mitigate this issue, the data captured at different angles is to be fused in order to achieve a global joint representation of the data that may be fed into the classification algorithm. 
For this, we have experimented with rotation-invariant representations of the data, e.g. by applying principal component analysis on the data~\cite{jolliffe2005principal} and also with mapping the data into alternative higher dimensional representations~\cite{li2021rotation}. 
In particular, L1-norm maximizing algorithms, such as~\cite{kwak2008principal,nie2011robust} generally achieve good results (L2-norm based algorithms are more susceptible to noise and outliers in the point cloud data). 
However, we observed that, for the data set we utilized, the large number of points reflected by the torso of the person, and which is similar for all gestures conducted, causes significant confusion among classes when applying PCA to achieve rotational invariance. 
While we are currently working on a solution for this problem, we have, for this article, instead implemented the feature fusion within a neural network architecture as described in section~\ref{sec:angleDependency} and in section~\ref{sec:rotationResilience}. When multiple UEs collaborate by jointly sensing the same situation, the neural network can be also trained, or updated periodically, using a distributed approach that leverages the NR-sidelink, namely the Federated Learning, described in section~\ref{sec:learning_processing}.  

\section{Instrumentation}
\label{sec:instrumentation}

\begin{figure}
    \centering
    \begin{subfigure}[b]{0.48\columnwidth}
       \includegraphics[width=\linewidth]{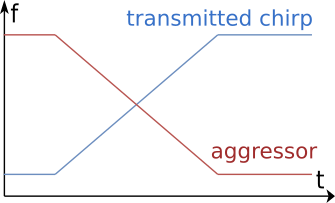}
       \caption{Crossing chrip interference.}
       \label{fig:interference_crossing} 
    \end{subfigure}
    \begin{subfigure}[b]{0.48\columnwidth}
       \includegraphics[width=\linewidth]{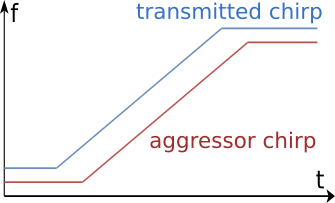}
       \caption{Parallel chirp interference.}
       \label{fig:interference_parallel}
    \end{subfigure}
    \caption{Crossing and parallel chirp interference between transmitted chirp (victim) and aggressor chirp.}
    \label{fig:interference}
\end{figure}

Although using multiple NR sidelink-based radars for sensing the environment can potentially address the shadowing effect and the low-resolution of the radar in z-axis \cite{palipana2021pantomime}, it also gives rise to issues due to the interference between radars including noise floor degradation, blind-spots at certain ranges or directions, as well as ghost objects. 
In this section, we discuss two different types of interference that can occur with FMCW radars: 
crossing chirp interference and parallel chirp interference. 
As shown in Fig. \ref{fig:interference}.a, crossing interference occurs when one chirp (referred to as aggressor in the following) crosses the chirp of another chirp (referred to as victim, since it is falling victim to the interference of the aggressor). 
This type of interference typically increases the noise floor resulting in a reduction in Signal to Noise Ratio (SNR) of the real targets thereby affecting detection and creating momentary blind-spots. The glitch duration in crossing interference is given by:

\begin{equation}
    glitch\_duration = \frac{bandwidth}{|slope_{aggressor}-slope_{victim}|}
    \label{eq:glitch_duration}
\end{equation}

According to Equation \ref{eq:glitch_duration}, the glitch duration for two crossing interferers is typically low and affects few samples.

The parallel interference is shown in Fig. \ref{fig:interference}.b. This type of interference occurs when the aggressor chirp and the victim chirp have the same slope. If the delay in the start of a chirp between different radars is within microsecond, the aggressor chirp will be within the bandwidth of the entire chirp of the victim. 
This type of interference results in ghost objects at random distance with random velocity that do not exist in the environment but are detected by the radar. 
Since such interference will occur only when the NR sidelink operating radars start nearly simultaneously, the probability of it is small. 
During the experiments we employed the built-in capability of the radar in interference detection to avoid any interference issues in the recorded dataset.

\subsection{Addressing angle dependency in NR sidelink based RF-sensing}
\label{sec:angleDependency}
\begin{figure*}
    \centering
    \includegraphics[height=7cm]{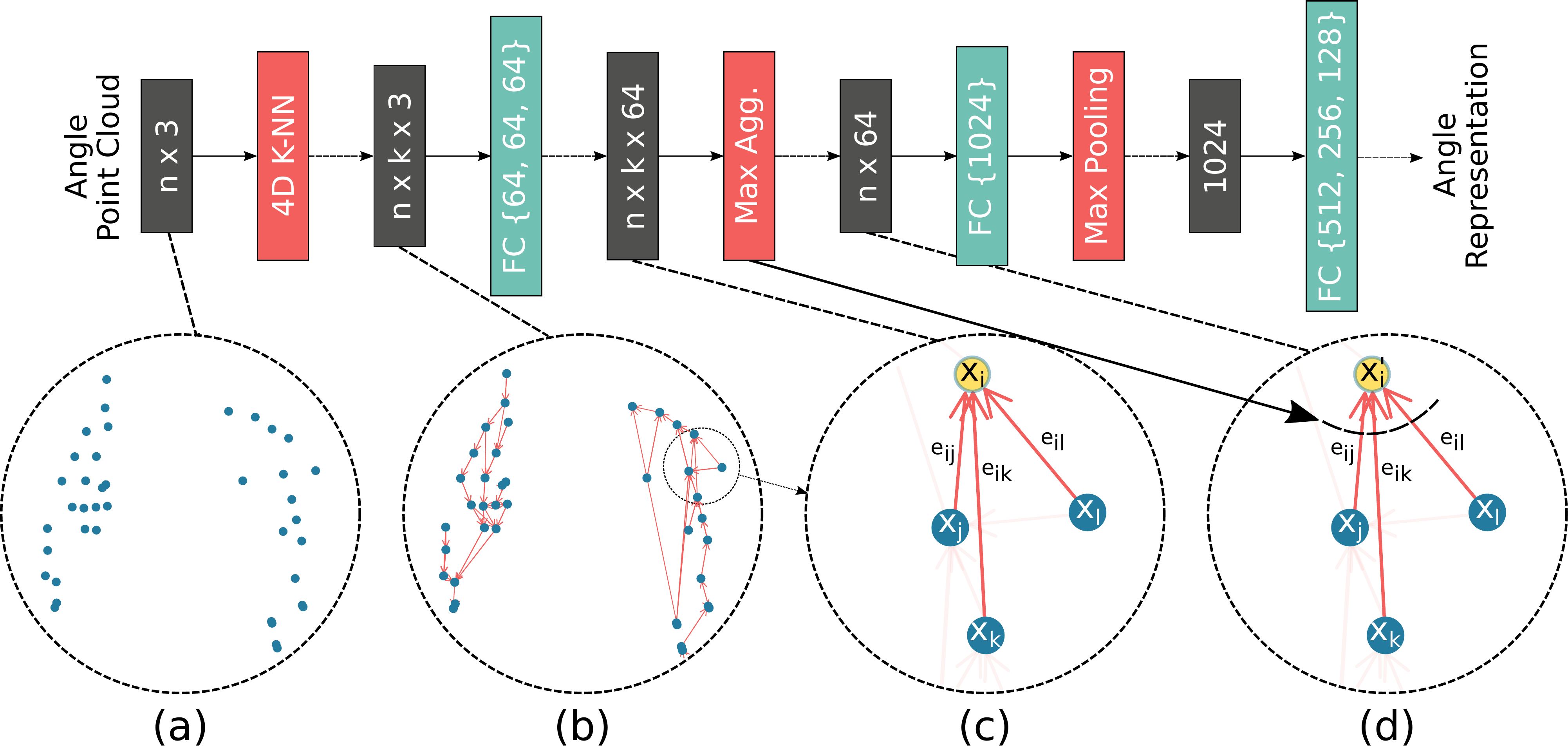}
    \vspace*{0.3mm}
    \caption{Schematic of the proposed encoder. (a) Point cloud generated by the radar for a gesture; (b) The graph representation of the gesture after applying the proposed K-NN to reflect the temporal dependency; (c) Edge features for each incident edge at a central point $i$ are calculated in this step; (d) The representation for all the nodes are calculated by applying an aggregation function over the edge features}
    \label{fig:encoder_arch}
\end{figure*}

\begin{figure*}
    \centering
    \begin{subfigure}[b]{.98\columnwidth}
       \includegraphics[width=.98\linewidth]{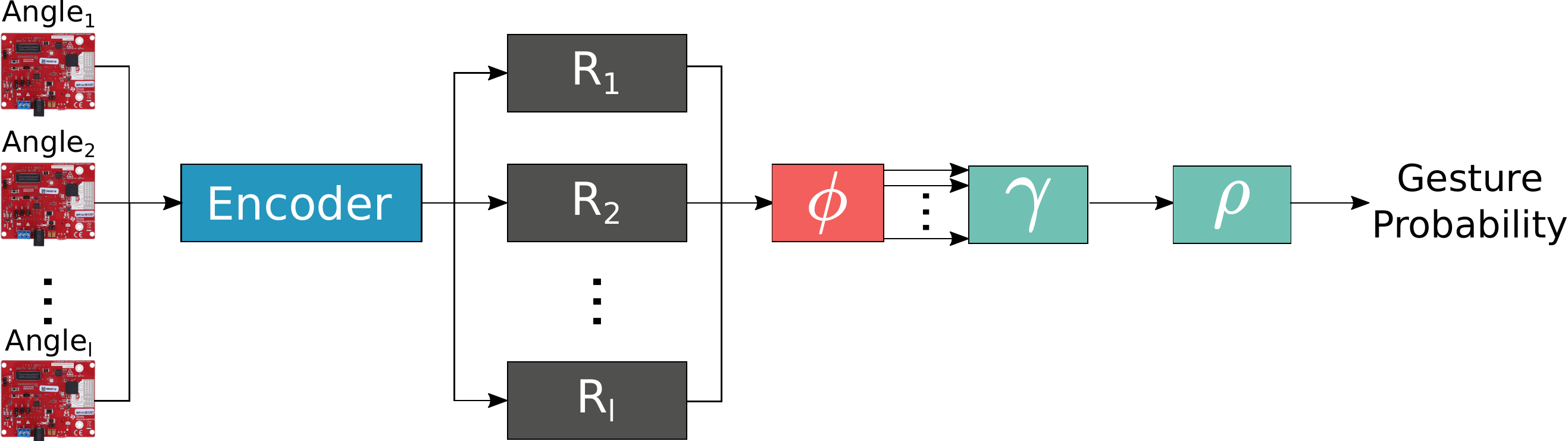}
       \caption{Each angle representation vector $R$ is fed into the same $\phi$ function; outputs are aggregated through different pooling mechanisms}
       \label{fig:Ng2}
    \end{subfigure}
    \begin{subfigure}[b]{.98\columnwidth}
       \includegraphics[width=.98\linewidth]{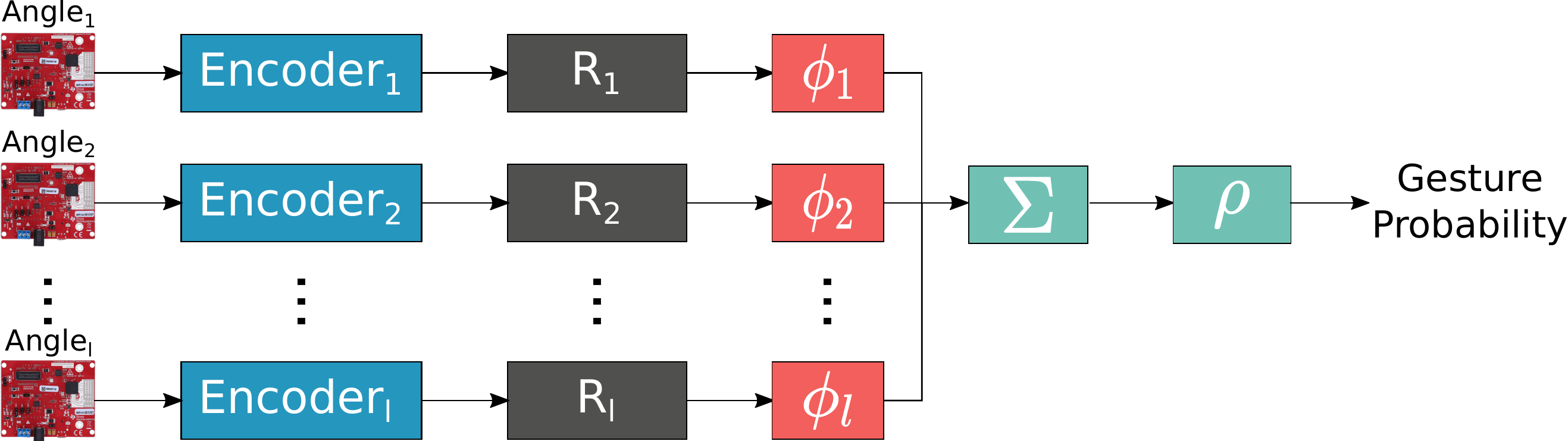}
       \caption{Different $\phi$ functions are applied on each angle representation $R$; output probability is the normalized sum of the per class score}
       \label{fig:Ng1} 
    \end{subfigure}
    \caption{Schematic view of the two proposed approaches: (a) orientation independent and (b) orientation tracking}
    \label{fig:gest_classi}
\end{figure*}
For each angle $a$ in the set of all angles of the NR-sidelink based radars, 
a motion point cloud $X_a=\{x_1,\dots, x_n\}$ is available in which each point has a feature set of $x_i=\{f_i^1, ..., f_i^F\}$. 
For each of the points in $X_a$, a frame number $f_i^s\in x_i$ is part of this feature set which defines the temporal aspect of the motion point cloud. 
To process the motion point cloud corresponding to a radar, a graph is generated and processed to obtain the representation of a gesture for each angle. 
To capture the temporal evolution of the gesture, the graph is generated over the nearest neighbours in temporal (frame-wise) dimension so that each point is connected to the closest points in the Euclidean space from the previous frames. 
To achieve this, a masking scheme is utilized in which the masked set of points $\mathcal{F}_{x_i}$ for $x_i$ is obtained through:
\begin{equation}
    \mathcal{F}_{x_i} =
    \{x_j : \forall x_j \in X, f_j^s > f_i^s\}.
    \label{eq:knn_points_from_frame}
\end{equation}

Moreover, the distance between two points is defined as the Euclidean distance of all the corresponding features of points including $f_i^s$ and is obtained through:

\begin{equation}
    D_{x_i,x_j} =
    \begin{cases}
    ||x_i - x_j|| : x_i,x_j \in X_a ,& \text{if } x_j \notin \mathcal{F}_{x_i},\\
    \infty ,              & \text{otherwise},
    \end{cases}
    \label{eq:knn_distance}
\end{equation}

where $D_{x_i,x_j}$ and $||.||$ denote the distance between $x_i$ and $x_j$ and the L$_2$ norm, respectively.
Finally, for each point K nearest points are chosen as neighbours in the graph according to $D_{x_i,x_j}$, resulting in a Temporal graph $\mathcal{G}_a=(\mathcal{V}_a,\mathcal{E}_a)$ where $\mathcal{E} \subseteq X_a \times X_a$ is the set of directed edges between each point and its neighbours.

To process the generated graph, an Edge Convolution Network (ECN) \cite{whang2018DEC} is applied and a vector of gesture representations is gathered. 
In the ECN, each convolution layer follows a message passing algorithm, in which features of each point in the graph gradually propagate to its neighbours as messages and the incoming messages for each point are aggregated with the features of the point itself using learnable, differentiable functions. 
In this network, in each convolution layer, the hidden representation of each point is updated through:
\begin{equation}
    \begin{aligned}
        h_{i}^0 &= x_i \setminus \{f_i^s\},\\
        h_{i}^l &=\underset{j:(i,j)\in \mathcal{E}}{\Gamma}M_\theta(h_i, h_j-h_i),
    \end{aligned}
    \label{eq:dynamic_edge_convolution}
\end{equation}

in which, $h_i^l$ is the hidden representation of point $i$ in convolution layer $l$,
$\setminus$ is the set subtraction operator, message function $M_{\theta}:\mathbb{R}^F \times \mathbb{R}^F \rightarrow \mathbb{R}^{F'}$ is an Multi Layer Perceptron function with learnable weight set of $\theta$, $\Gamma$ is a channel-wise symmetric max pooling function \cite{GoodBengCour16} applied on the messages of the edge emanating from each neighbor. 
To obtain the representation vector of each angle $a$ motion point cloud $\mathcal{R}_a$, a symmetric max pooling function of $\Gamma$ is applied on the hidden representation of all points in $X_a$ as:
\begin{equation}
        \mathcal{R}_a = \underset{i\in \mathcal{V}}{\Gamma}h_i.
    \label{eq:angle_representation}
\end{equation}

\subsection{Rotation-resilient Classification of gestures}
\label{sec:rotationResilience}
The set of all angle motion point cloud representation vectors is denoted by $\mathbb{G} = \bigcup_{a\in \mathcal{A}}R_a$. 
In order to predict the gesture from $\mathbb{G}$, two different approaches of angle invariant and user orientation tracking are considered. 
The pipeline of each is demonstrated in Fig. \ref{fig:gest_classi}. 
In the first approach (Fig. \ref{fig:gest_classi}(a)), a set of permutation invariant functions with respect to elements of $R_a$ are used to process $\mathbb{G}$ and to predict the gesture label. 
In the second method (Fig. \ref{fig:gest_classi}(b)), for each vector, a set of learnable functions is considered. 
The details of each method are presented in the following.

\subsubsection{Angle invariant prediction}
For this approach, each $\mathcal{R}_a$ is fed into a function that calculates the embedding of the vector independent of others and the output of embeddings are aggregated through a pooling mechanism $\gamma$. This method is commonly used to process sets of vectors \cite{lee2018Settransform} and is achieved via
\begin{equation}
    P = \rho[\gamma(\phi(R_1), \phi(R_2),\dots, \phi(R_m))].\label{eqn:P}
\end{equation}
In equation~\ref{eqn:P}, $P$ is the probability of each class, $\rho$ and $\phi$ are learnable functions. For our method, we employ 1-dimensional convolution layers for  $\phi$ and MLPs followed by a Softmax function for $\rho$.
For the pooling operator, we provide a choice between different pooling methods of \textbf{Max pool} \cite{GoodBengCour16}, \textbf{Attention pool} \cite{lee2018Settransform}, and element wise summing referred to as \textbf{Vote pool}.
Each of the pooling operators result in a different prediction model that has different performances when it comes to processing the gesture set $\mathbb{G}$. As opposed to the traditional maxpooling method which does not contain any learnable weights, the Attention pooling operator, aggregates the angle representation vectors into a single vector using learnable differentiable weights that optimizes the final gesture representation through the multi-head self-attention mechanism \cite{vaswani2017attention}. 
In the case of Vote pool, $\phi$ is replaced with a multi layered perceptron function for which the size of the last layer is equal to the number of gesture classes and $\rho$ is the Softmax function. 
The advantage of this method is that we do not need to keep track of the orientation of the user with respect to NR sidelink based radars in case the user changes its position with respect to the surrounding sidelink operating radars. 
\subsubsection{User orientation tracking prediction}
In this method, each $\mathcal{R}_a$ is fed into a separate function to embed calculation which are specific to each angle $a$. 
The output of each function is then a separate prediction for the label of a gesture. 
The final prediction of the gesture is obtained through a voting mechanism between all the function outputs. 
This method follows the formula
\begin{equation}
    P = \rho[\Sigma(\phi_1(R_1), \phi_2(R_2),\dots, \phi_m(R_m)]. \label{eqn:Prho}
\end{equation}
In equation~\ref{eqn:Prho}, $\Sigma$ is an element-wise summation function and $\rho$ is the normalization function for probability calculation. In this approach, we need to know which radar is located at which angle with respect to the user to feed the input motion point cloud to the corresponding $\phi_a$. For the $\phi$ functions we employ MLPs followed by a Softmax function to predict each class probability. 

\section{Learning, processing and data fusion in distributed scenarios}
\label{sec:learning_processing}
In this section, distributed processing and machine learning (ML) tools are proposed to leverage the sidelink resources for rotation-resilient gesture classification. In particular, we assume that the radars are equipped with a low-power processing unit that supports gradient-based neural network model optimization~\cite{kairouz2021advances} as well as real-time data fusion and classification. Radars can be randomly distributed in the space however, to simplify the reasoning, we assume that these are deployed according to a ring topology as depicted in Fig~\ref{fig:setupA}.

Conventional machine learning systems fuse and process the motion point clouds of each radar on a data center, typically colocated with the gNB. 
However, such centralized processing compromises data privacy, lacks scalability and often requires an intensive use of the uplink radio channel, for moving raw datasets. 
A distributed policy based on a federated learning (Federated Learning) tool is explored that leverages device-to-device sidelink communications.
Federated Learning~\cite{kairouz2021advances} is a recently proposed distributed machine learning paradigm that allows the parameters of a Machine Learning model to be collectively optimized across several resource-constrained
wireless devices~\cite{FL1} equipped with low-power tensor processing units. 
The proposed Federated Learning system lets the radar devices act as local learners, beside point cloud data producers. 
The learnable model parameters described in Section~\ref{sec:instrumentation} (the MLP function $\theta$ and the classifier $\phi(.)$) are first optimized locally from training data. 
Next, the radars mutually exchange the local models using the sidelink resources. 
The NR sidelink operating radars in-turn improve the local parameters by fusing the received contributions. 
This procedure continues for a new optimization round until the model satisfies a target accuracy using validation data. 
Federated Learning is privacy-preserving by design, as it keeps the raw motion point clouds $X_{a}$ on the radar devices~\cite{kairouz2021advances}. It also unloads the uplink cellular link and obviates the need for a data center.

Targeting rotation-resilient classification of gestures, some considerations on Federated Learning implementation are required. 
In user orientation tracking, 
the learned functions are location specific, namely specific to the radar relative position (angle $a$): each radar thus autonomously learns the local angle representation and the classifier parameters $\phi_{i}(.)$ using training data.  
For angle invariant prediction, 
the model parameters $\phi(.)$ and $\gamma$ in eq. (\ref{eqn:P}) are learned collectively via Federated Learning: 
on each round, the fusion of the received models can be orchestrated by the initiating UE and implemented via distributed weighted averaging~\cite{FL2}. Federated Learning can be used both for initial training of model parameters and for periodic retraining or update. In what follows, considering the complexity of the model, we adopt the latter approach.


\section{Evaluation}
In this section, we evaluate the performance of the proposed models in terms of recognition accuracy under various conditions. 
The evaluation metrics we use are \textit{balanced average accuracy} and \textit{Area Under the ROC Curve (AUC)}.
Balanced average accuracy is defined as the average of the recall obtained on each class. 
AUC is equal to the probability that a classifier will rank a randomly chosen positive instance higher than a randomly chosen negative one which in turn quantifies the discriminatory power of the classifier. 
\subsection{Data collection}
\begin{figure*}
    \centering
    \begin{subfigure}[b]{\columnwidth}
        \centering
       \includegraphics[width=.6\linewidth]{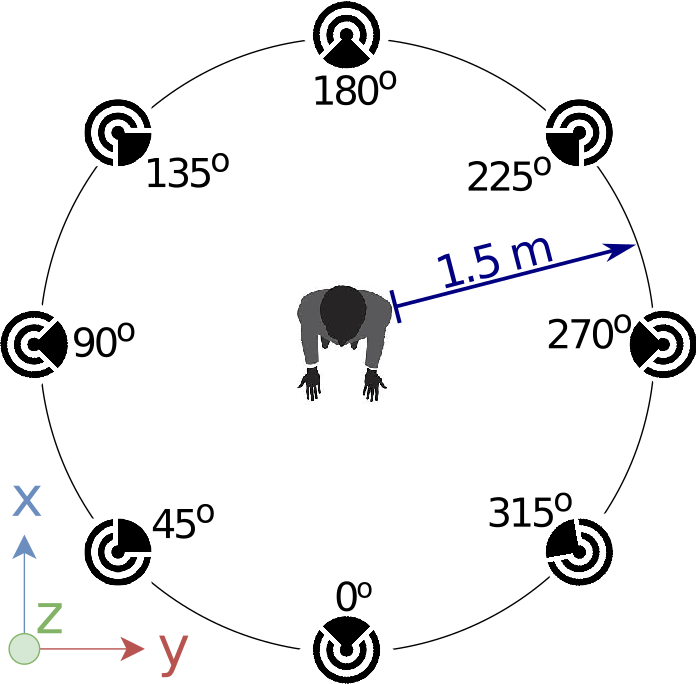}
       \caption{positioning of the radars with respect to the participant}
       \label{fig:setupA}
    \end{subfigure}
    \begin{subfigure}[b]{\columnwidth}
       \includegraphics[width=\linewidth]{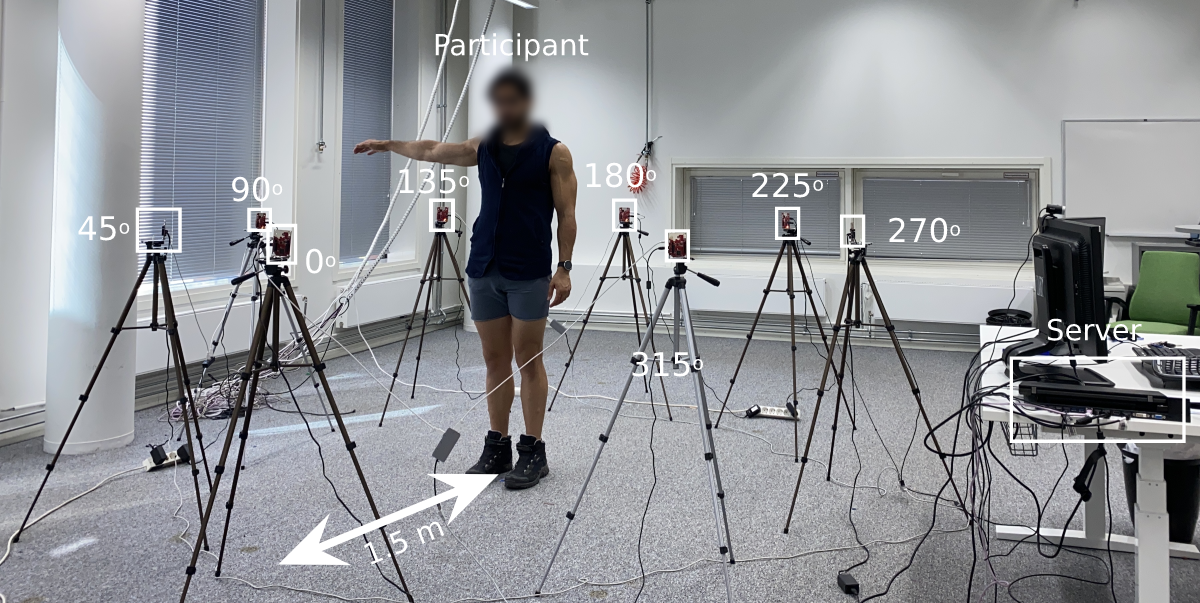}
       \caption{actual environment with a participant performing a gesture and radars recording}
       \label{fig:setupB}
    \end{subfigure}
    \caption{Experimental setup for data collection.}
    \label{fig:setup}
\end{figure*}
%

We utilize mmWave FMCW radars i.e. IWR1443 for the experiments. These radars transmit a sinusoidal signal with linearly increasing frequency, called chirp, and process the reflected signal to extract information like range, velocity, and the angle of the targets in the environment. 
The starting frequency of the chirps is 77 GHz and the final, highest frequency is 81 GHz, resulting in a 4~GHz bandwidth. 
The evaluation kit produces point clouds from dense tensors through a signal processing pipeline~\cite{palipana2021pantomime}.

Following the same gesture set in~\cite{palipana2021pantomime}, we collected 21~classes of gestures shown in Fig.~\ref{fig:gesture_set} from 15 subjects repeating each gesture 10~times. 
We installed 8~radars around the participant in a 1.5m distance. 
The positioning of the radars with respect to the participant is shown in Fig.~\ref{fig:setupA} and Fig.~\ref{fig:setupB}. 
We recruited 15 participants (7 female) aged 19–50, weight 50–105 kg, and height 1.55–2 m for data collection. 
Participants were unpaid volunteers recruited via mailing lists, external advertisements, and flyers distributed among several of our university’s departments. Before starting the data collection process, the experimenter demonstrated the correct way of performing gestures through pre-recorded videos. 
Moreover, while recording, the experimenter verified gestures by observing a visualized point cloud on the screen. In case the gesture was performed incorrectly, the participant was asked to repeat the gesture. 
To collect the data from different angles, as shown in Fig.~\ref{fig:setupB} we connected 8 radars to a laptop through USB ports.
The final dataset consists of 25,200 gesture samples.

To receive the point cloud from each radar and save them to files, we developed an open source software called \textit{mmWave Data Collector} which supports multiple IWR1443 and AWR1642 radars. 
\textit{mmWave Data Collector} uses Robot Operating System (ROS)\footnote{https://www.ros.org/} extending its functionalities to support more than one sensor simultaneously.

For the train/test data split we follow a population dependent cross validation strategy to answer the question that how well the models perform recognizing gestures from unseen users. We use 10 users for train/validation and 3 users for inference purpose.

\subsection{Data Preprocessing}
Since we record each gesture for 3 seconds, they have different different number of frames. However, the baseline models require a fixed number of frames and points in each frame. After receiving moving point clouds from the radar, we divide the points in such a way that we have the same number of frames.
After this, we perform point re-sampling on each frame to fix the number of points in frames for all samples. 
To fix the number of frames, we empirically distribute the points into 32~frames (for a point cloud with $n$ points, the first $n/32$ points are considered as the first frame, the second $n/32$ are considered as the second frame, and so on).
Moreover, to fix the number of points in each frame while preserving the spatial structure, we utilize the density-based re-sampling method introduced in~\cite{cohen2006learning}. 
Assume $n/32$ as the desired number of points in each frame. 
To down-sample the points, $\mathcal{K}$-means algorithm is used by setting $\mathcal{K}$ equal to $n/32$ and selecting the centroids of the clusters as the points in the frame. 
To up-sample the points, Agglomerative Hierarchical Clustering (AHC) is applied iteratively and the centroids of the clusters are added as new points to the frame until we have the desired number of points. 
We empirically set the number of points in each frame to~32.

\subsection{Data Augmentation}
To increase the generalizability of the models (see section~\ref{sec:angleDependency}), we apply five data augmentation techniques on each batch during the training phase. 
Random translation of the gesture up-to 10cm, random scaling between 0.8 and 1.25, random jittering (point-wise translation) based on a Gaussian distribution with $\mu=0$ and $\sigma=0.01$, random clipping of 0.03m, and random shuffling of points in each frame to preserve the spatial and temporal dependencies while changing the input representation are the five data augmentation techniques applied on-the-fly.

\subsection{Model Implementation}
The proposed models are implemented using PyTorch~\cite{paszke2019pytorch} and PyTorch Geometric~\cite{fey2019fast}. 
For initial training of the models, including baselines and the proposed models except for Pantomime and PointGest, we used a server with 64GB of RAM and a Tesla V100 16GB GPU. Federated Learning, discussed previously, can be also used to distribute the training tasks on the radar devices sharing the NR-sidelink. 
Since Pantomime and PointGest are computationally expensive, we used a Tesla V100 32GB GPU to train them.

During the training phase, we also utilize an early stopping strategy with a patience of 100 epochs to avoid over-fitting of the models. During training, if no improvement is observed on the validation set in terms of cost function within the patience period, the training is stopped and the best model is saved for the inference phase. We use negative log likelihood between softmax followed by logarithm of class scores and the ground-truth labels as the loss function~\cite{Goodfellow-et-al-2016}. Adam Optimizer~\cite{kingma2014adam} with a step-decay strategy to decrease learning rate is used to optimize the loss function:
\begin{equation}
    L_{r} = L_{init} \cdot d_{r} ^ {\lfloor \frac{e} { e_{r}}\rfloor} 
    \label{eq:step_decay}
\end{equation}
In equation~\ref{eq:step_decay}, $L_{r}$ is the learning rate used at each epoch, $L_{init}$ is the initial value of the learning rate, $d_{r}$ is the drop rate after every $e_{r}$ epochs, $e$ is the current epoch and $\lfloor\cdot\rfloor$ is the floor operator. In our setup $L_{init}$ is $0.001$, $d_{r}$ is $0.5$, and $e_{r}$ is $80$.

\subsection{Classification Results}
In Table~\ref{table:soa_comp}, the performance of our proposed models (Max Pool, Attention Pool, Vote Pool, Orientation Tracking) is compared to the baseline models PointNet++~\cite{qi2017pointnet2}, DEC~\cite{whang2018DEC}, PointLSTM~\cite{min2020efficient}, PointGest~\cite{salami2020motion}, and Pantomime~\cite{palipana2021pantomime}. 
In PointNet++ and DEC, the data are combined through time and angle resulting in a single point cloud representing the whole gesture from different angles since they are designed to process static point clouds. 
For PointLSTM, PointGest, and Pantomime, the four dimensional (4D; x-y-z and time) point clouds from different angles are combined frame-wise resulting in a temporal 4D point cloud representing the gesture from different angles since the models are able to process 4D point clouds. 
In this section, we assume that all angles are available in the inference phase and the orientation of the target w.r.t. the radars is known. 
As illustrated in Table~\ref{table:soa_comp}, two of our models, attention pool and orientation tracking, outperform state-of-the-art in terms of both average accuracy and AUC. The orientation tracking approach achieves an accuracy and AUC of 100.

\begin{table}[t]
	\centering
	\begin{tabular}{l *6c}
		\toprule
		\textbf{Model}
		  & \textbf{Acc.}
		  & \textbf{AUC}
		\\ \midrule
		Pointnet++
		  & 82.51
		  & 98.21
		  \\
		\gls{dec}
		  & 96.20
		  & 99.99
		\\
		PointLSTM
		  & 96.67
		  & 99.95
		\\
		PointGest
		  & 95.45
		  & 99.93
		\\
		Pantomime
		  & 98.63
		  & 99.97
		\\
		
		\bottomrule
		\addlinespace[0.5em]
		Max pool(ours)
		  & 95.40
		  & 99.83
	    \\
		Attention pool(ours)
		  &  98.73
		  &  \textbf{100}
		\\
		Vote pool(ours)
		  &  98.41
		  &  99.73
		 \\
		Orientation tracking(ours)
		  &  \textbf{100}
		  &  \textbf{100}
		 
		\\ \bottomrule
		\addlinespace[0.5em]
	\end{tabular}
	\caption{Comparison with the state of the art on the collected dataset when all the angles are available in the inference phase.
	The best results per column are denoted in bold typeface.}
	\label{table:soa_comp}
\end{table}

\subsection{Angle Drop-out Results}
\label{sec:angle_dropout}
\begin{figure*}
 \includegraphics[width=\textwidth]{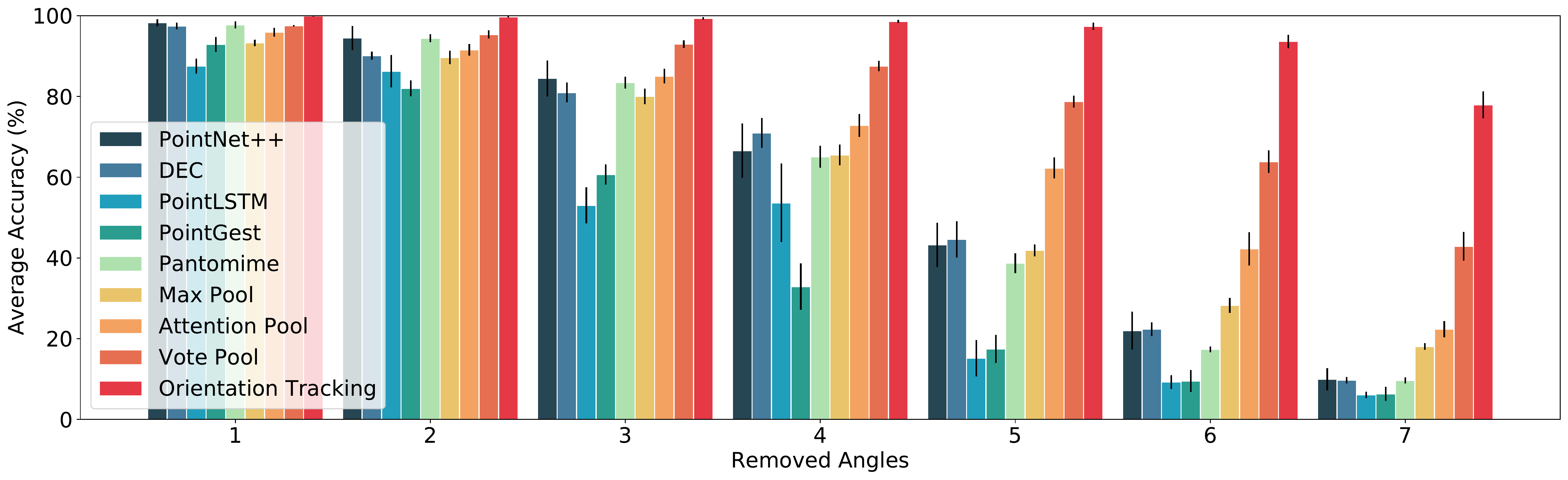}
 \caption{The performance of the proposed models compared to the baselines when not all the angles are available in the inference phase. Different number of angles are randomly removed in the inference phase.}
  \label{fig:dropout_angle}
\end{figure*}

In realistic scenarios, we might not have all angles available in the inference phase. 
In this section, we evaluate the performance of our proposed models as well as the baselines when different number of angles are available in the inference phase (from 1 to 7). 
For each batch (a set of 32 gestures), we randomly remove the specified number of angles and evaluate the performance of the models. 
We repeat each setting 10 times and report the average accuracy and standard deviation of each model in Fig.~\ref{fig:dropout_angle}. 
First, as we decrease the number of available angles, the standard deviation for all the models increases since not all angles have the same amount of information (see section~\ref{sec:angle_importance}). 
Second, the orientation tracking approach achieves a notable accuracy even for recognizing gestures using a single angle compared to the rest of the models. 
The drop in the accuracy for the orientation tracking model when we go from 8 angles to 1 angle is only 22\% while for rest of the models it is from 55\% to 90\%. Moreover, for extreme settings in which we have only one or two angles available, all the proposed models outperform the baselines since the baseline models are not capable of taking into account the angles.
This observation suggests that dedicating an encoder with an independent set of weights for each angle increases the generalizability of the model for settings with varying number of available radars (angles). 
However, this improvement comes with a cost of requiring knowledge of the target's orientation with respect to the NR sidelink operating radars. 

\subsection{Paired Angle Recognition Results}
\begin{figure*}
\setlength\tabcolsep{1.5pt}
\centering
\begin{tabular}{ccccc}
    \includegraphics[width=0.19\textwidth]{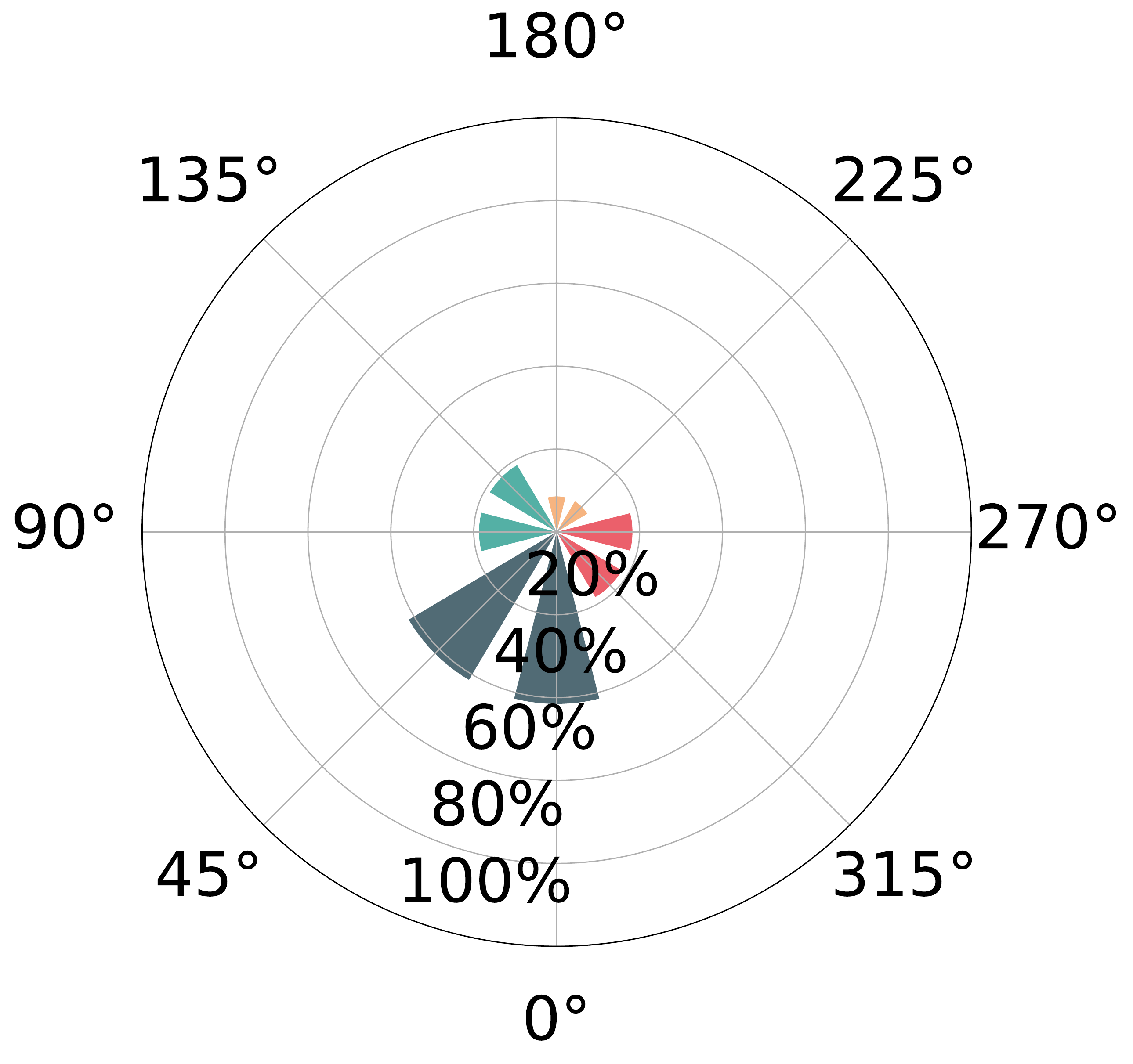} &   \includegraphics[width=0.19\textwidth]{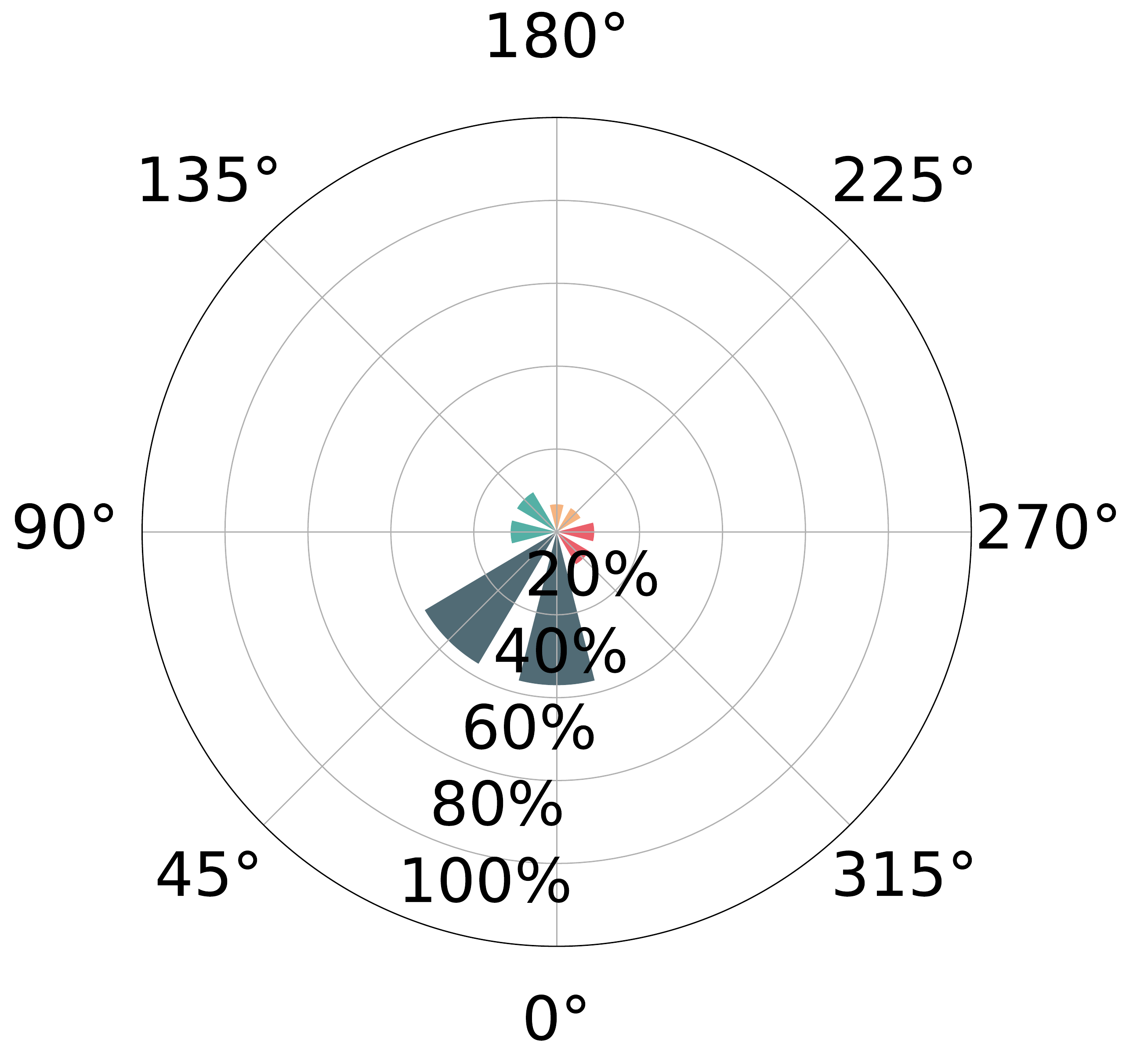} &  \includegraphics[width=0.19\textwidth]{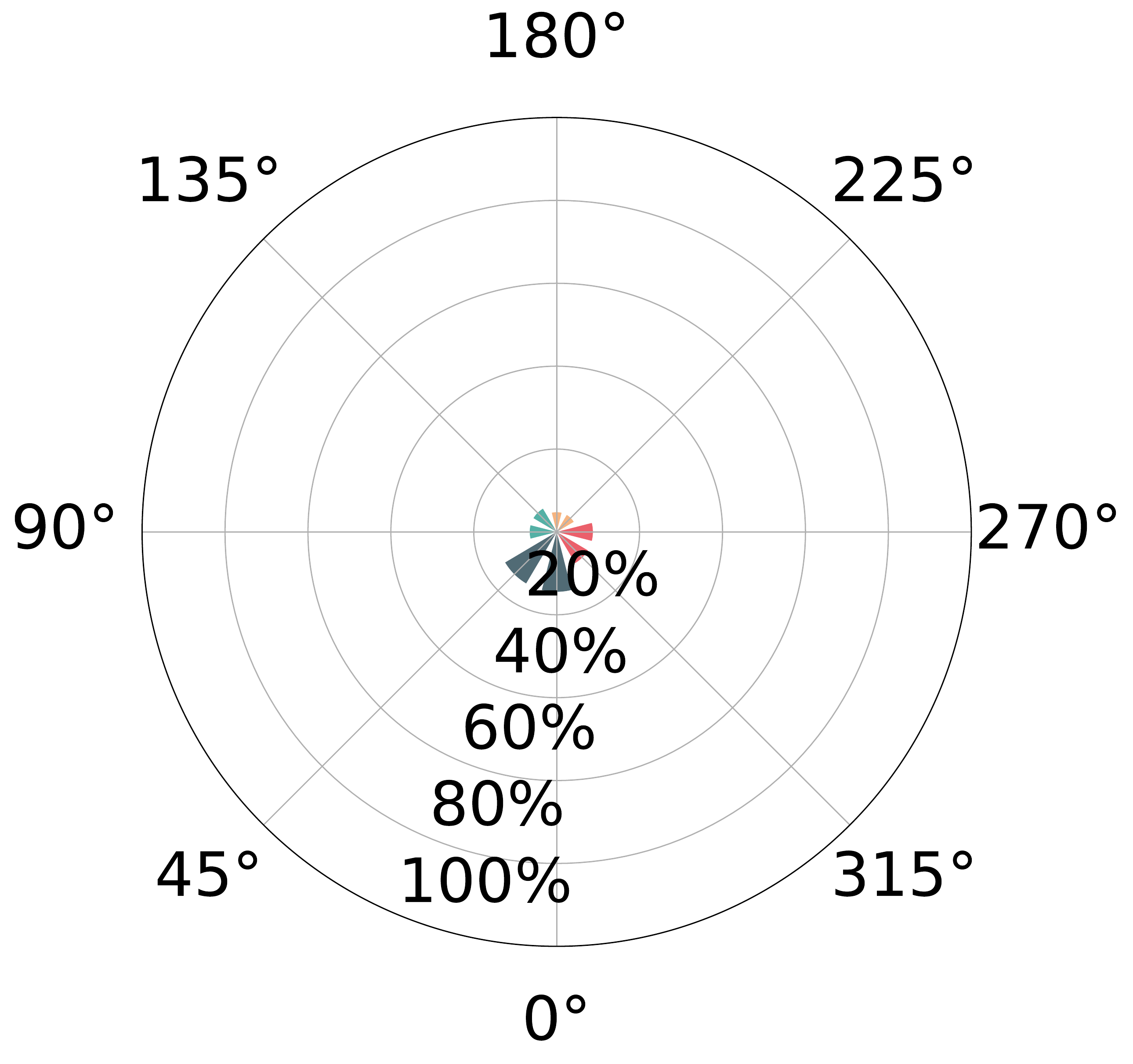} &   \includegraphics[width=0.19\textwidth]{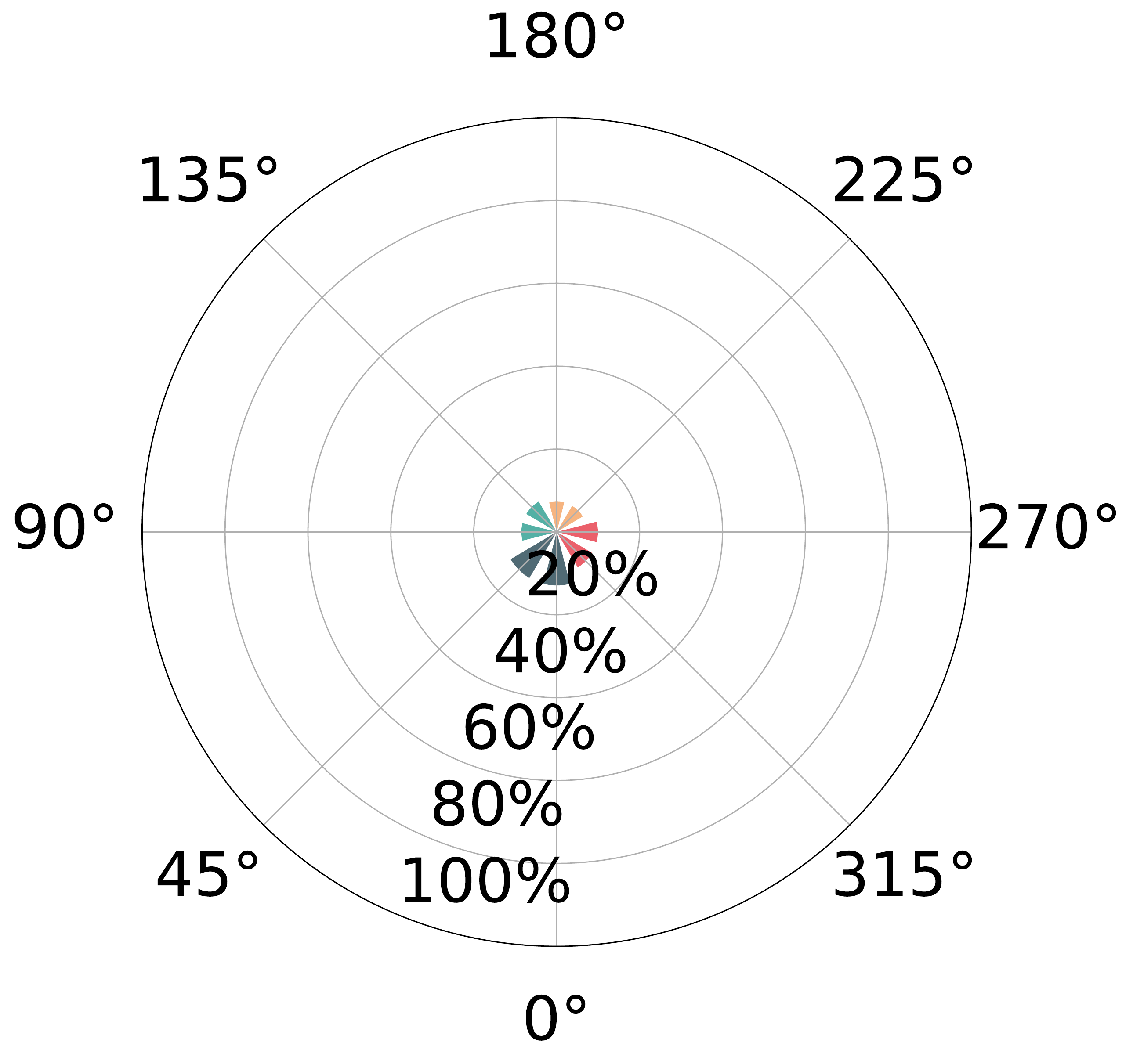} &   \includegraphics[width=0.19\textwidth]{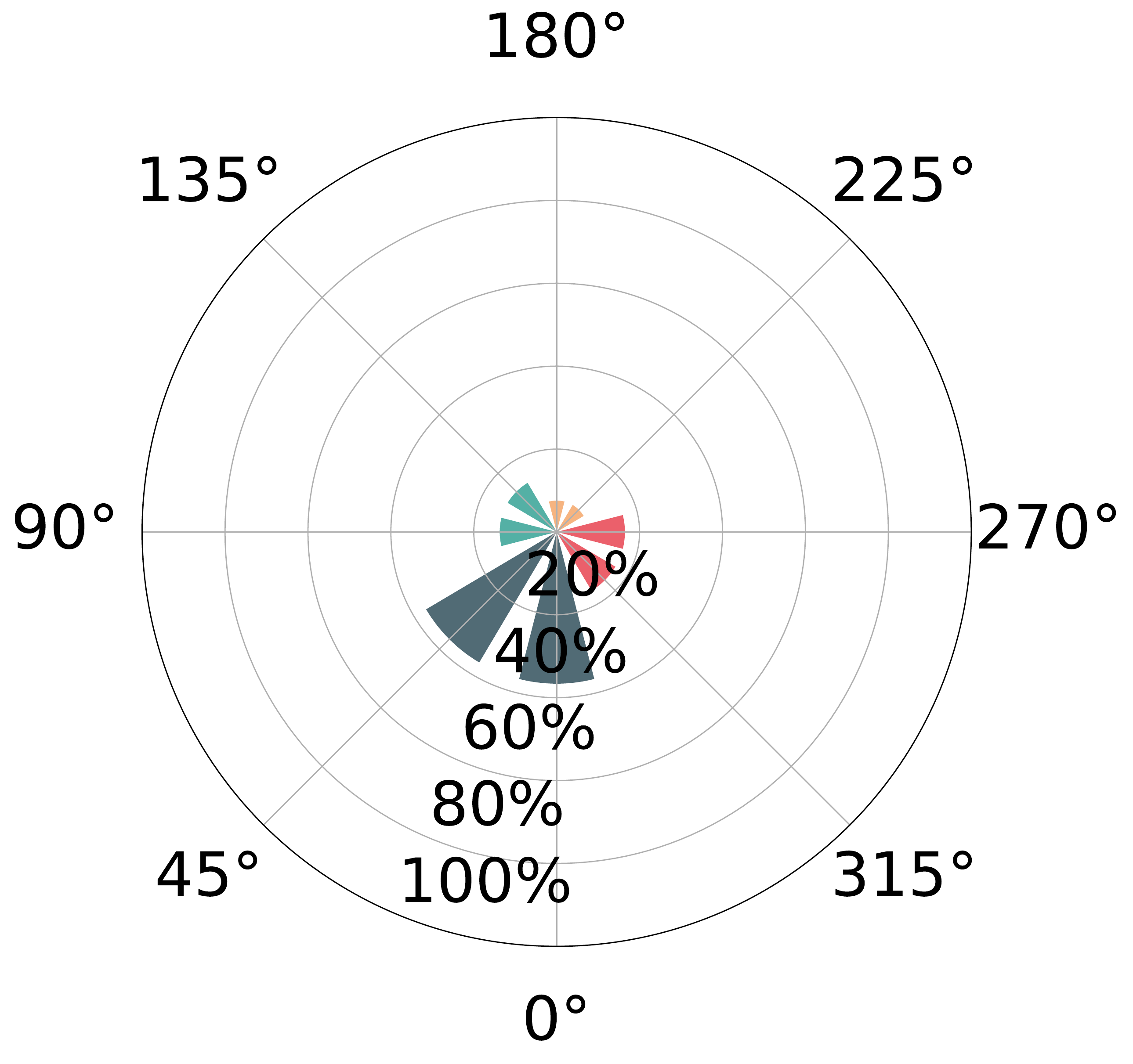} \\
    (a) PointNet++ & (b) DEC & (c) PointLSTM & (d) PointGest & (e) Pantomime \\
    \includegraphics[width=0.19\textwidth]{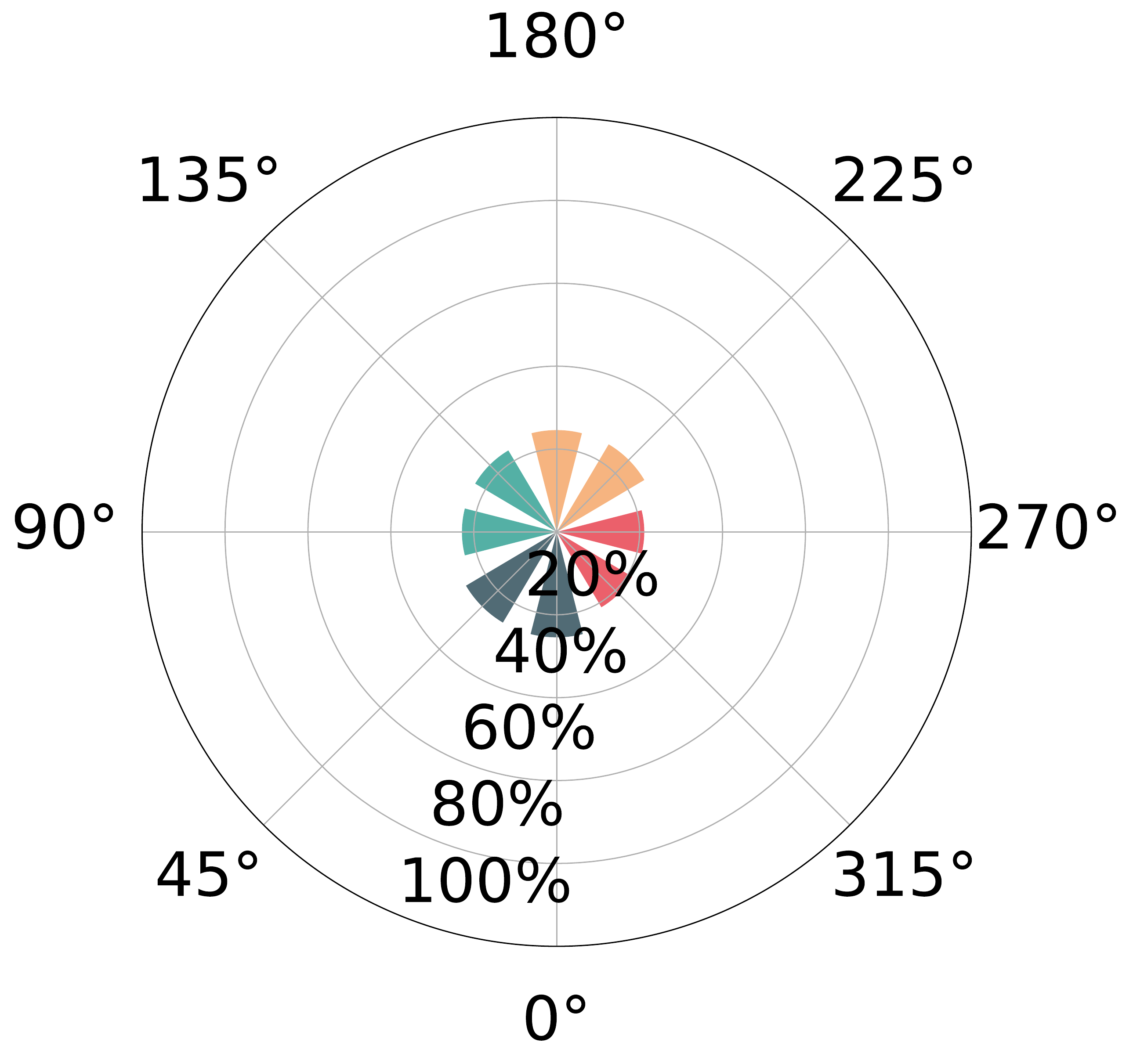} &   \includegraphics[width=0.19\textwidth]{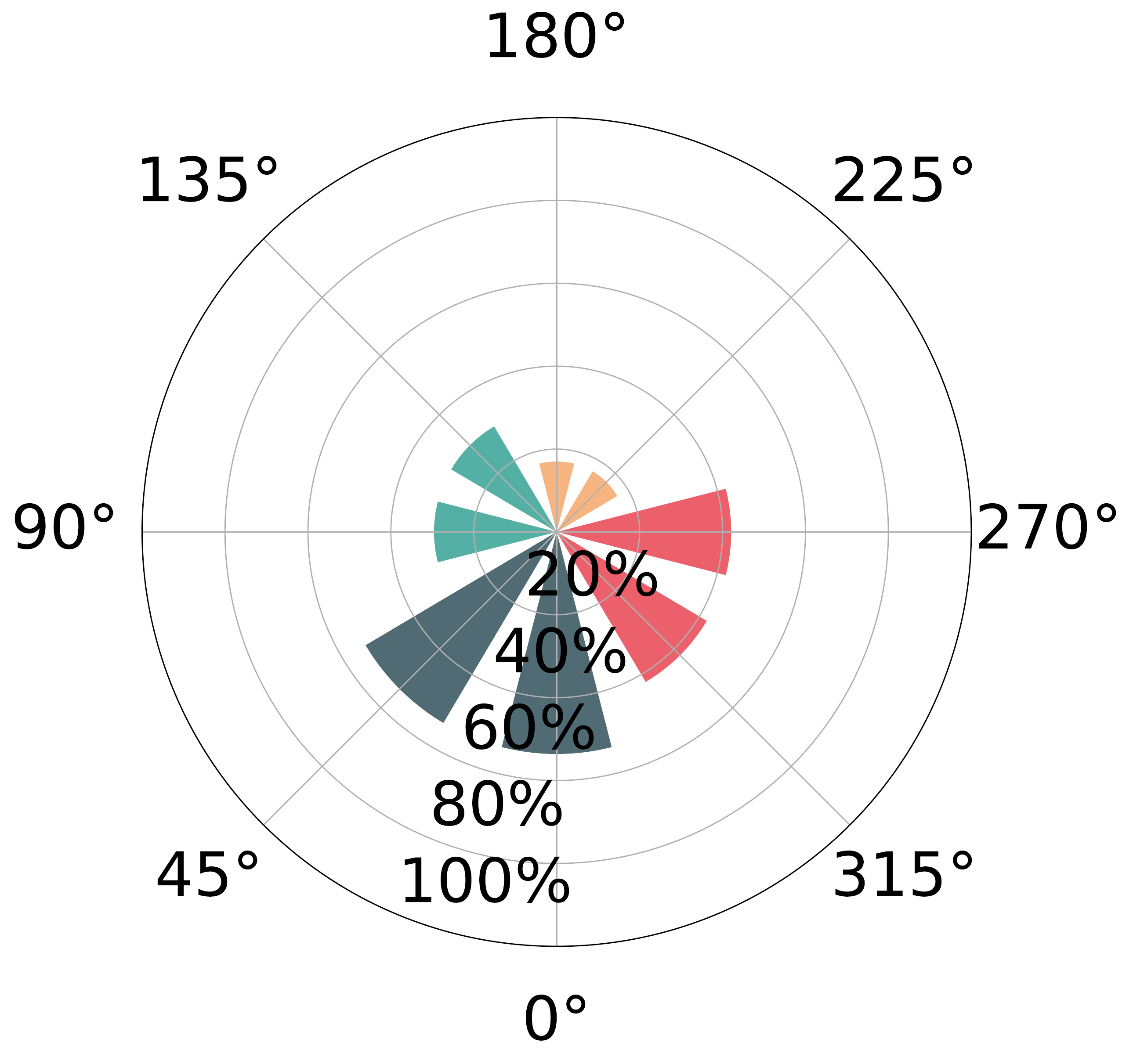} &  \includegraphics[width=0.19\textwidth]{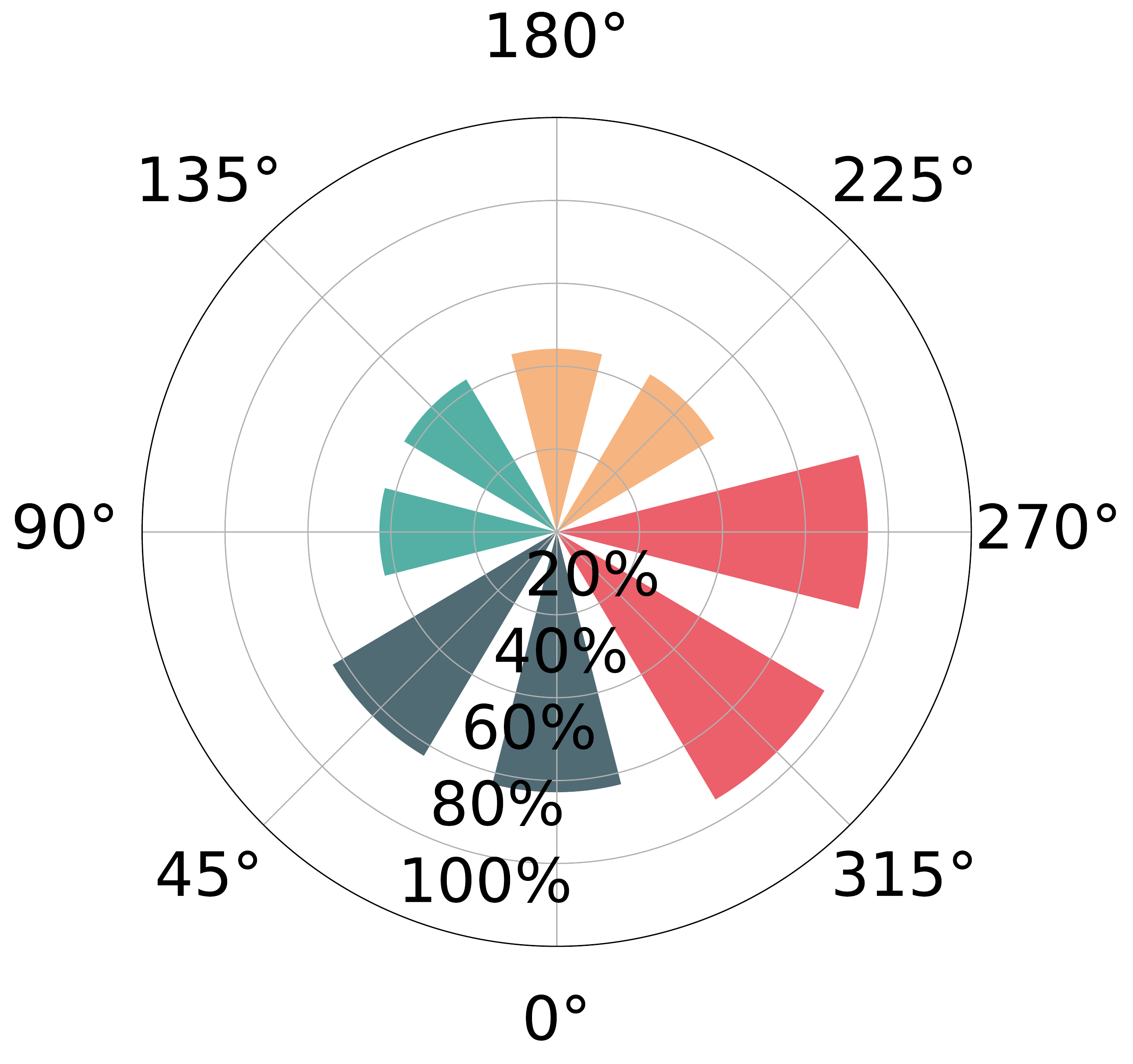} &   \includegraphics[width=0.19\textwidth]{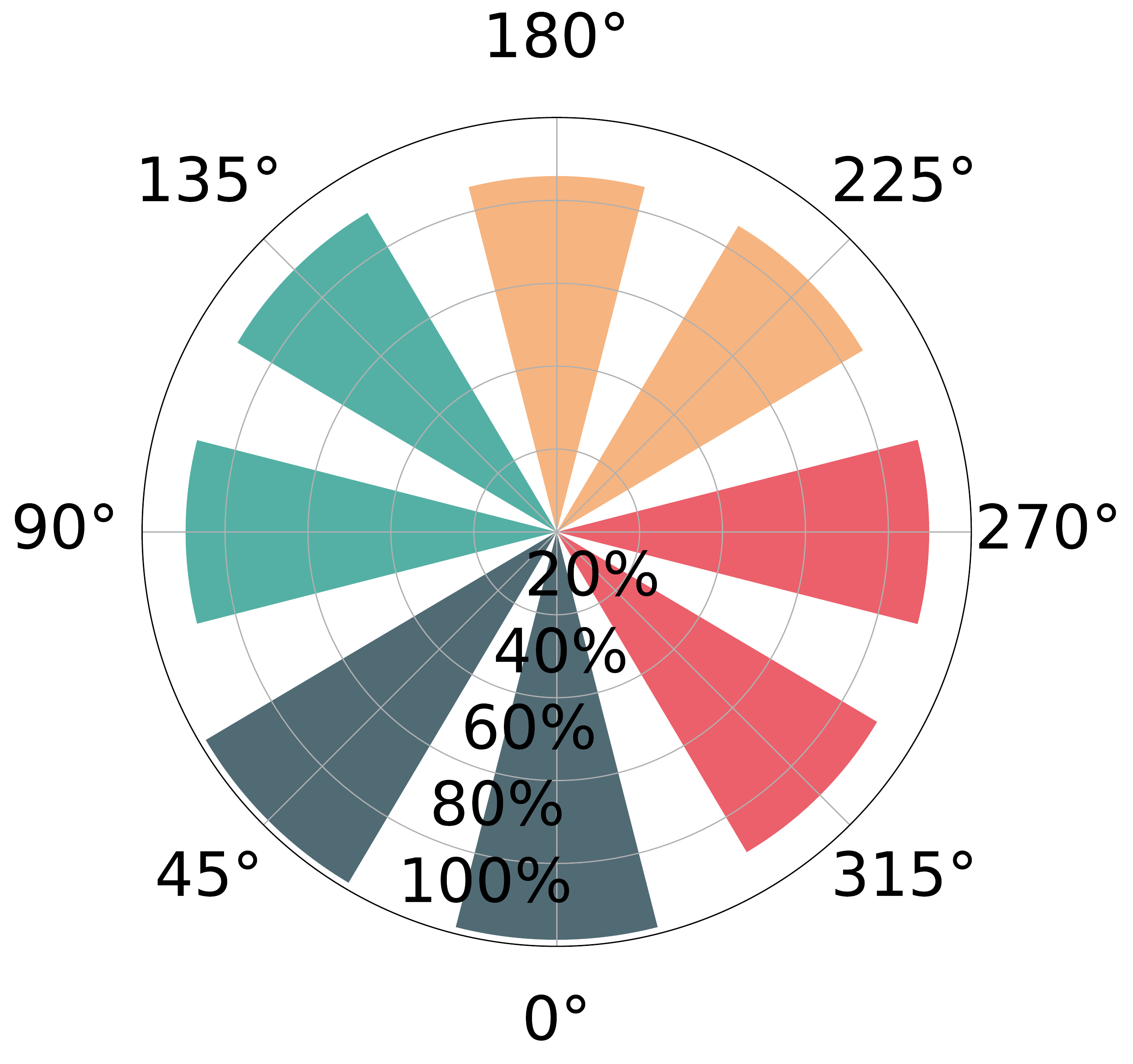} &   \includegraphics[width=0.19\textwidth]{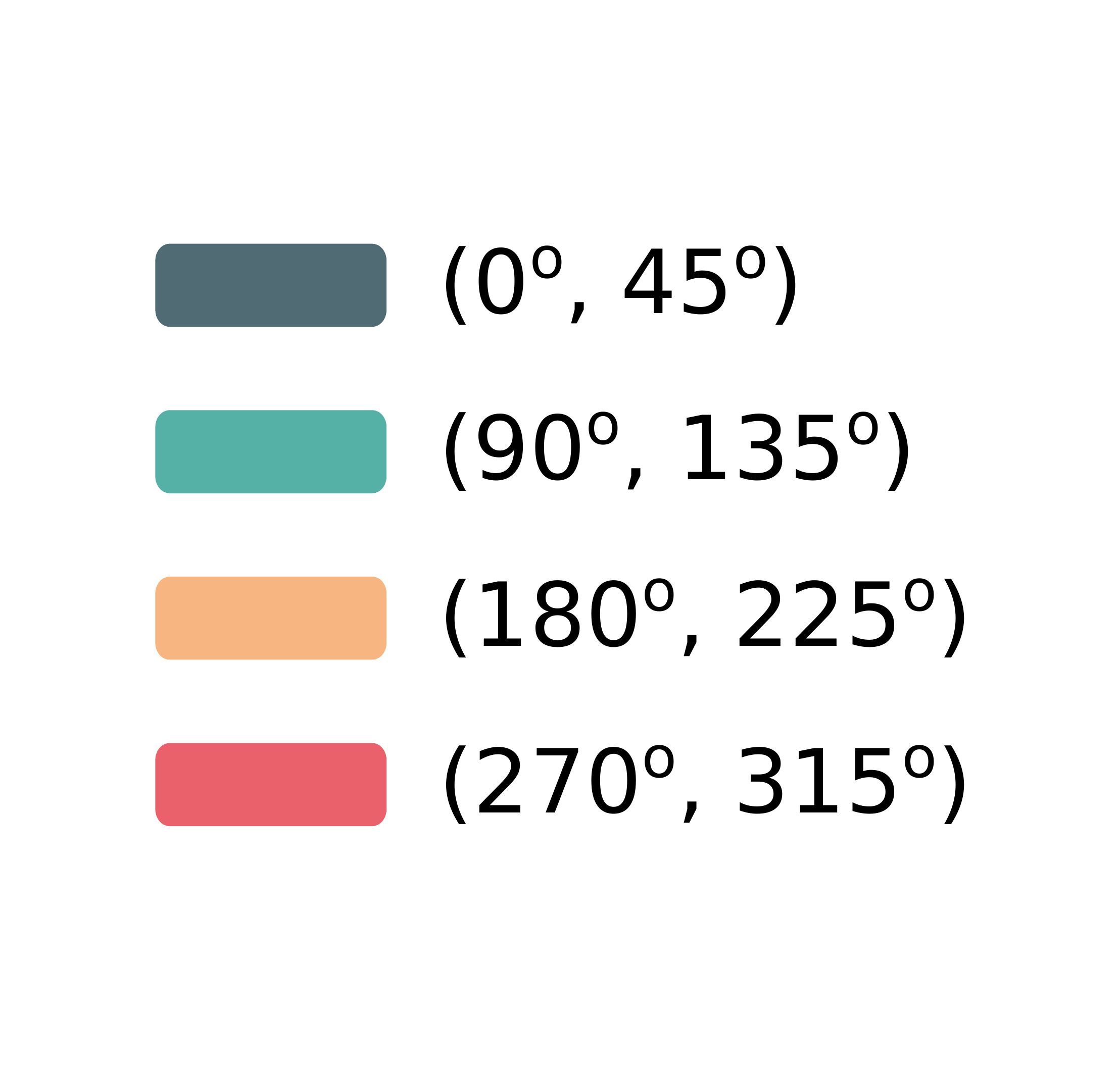} \\
    (f) Max Pool & (g) Attention Pool & (h) Vote Pool & (i) Orientation Tracking &\\
\end{tabular}
\caption{The performance of proposed models compared to the baseline models when we have different pair of angles in the inference phase. The metric shown in these plots is the average accuracy. For each model, we evaluate the performance for four different pairs of ($0^\circ$, $45^\circ$), ($90^\circ$, $135^\circ$), ($180^\circ$, $225^\circ$), and ($270^\circ$, $315^\circ$).}.
\label{fig:paired_angle_results}
\end{figure*}

In realistic scenarios, a person is not surrounded by 8~RF sensors as in our experimental setting and indeed, the actual angle of a sidelink operating radar with respect to the person performing the gesture might be random and arbitrary. 
This means that data is produced only by a few selected angles. 
For the recognition of gestures, we show that the orientation tracking approach is significantly better than the base line models (see section~\ref{sec:angle_dropout}). 
In particular, we pair angles, ($0^\circ$, $45^\circ$), ($90^\circ$, $135^\circ$), ($180^\circ$, $225^\circ$), and ($270^\circ$, $315^\circ$), to evaluate the models when data from only two deterministic angles are available. 
All models were trained on data from all angles while in the inference phase, the data was provided only from one of these angle pairs.
As shown in Fig.~\ref{fig:paired_angle_results}, the baseline models fail to generalize on a limited number of angles suggesting that they overfit the train data where all the angles are available. 
However, the proposed methods attention pool, vote pool, and orientation tracking outperform the baselines in all the four combination of angles. 
As we mentioned in section \ref{sec:angle_dropout}, the orientation tracking approach is resilient to the number of available angles in the inference phase suggesting that it is able to generalize on various number of angles.

\subsection{Angle Permutation Results}
\begin{figure}
 \includegraphics[width=\columnwidth]{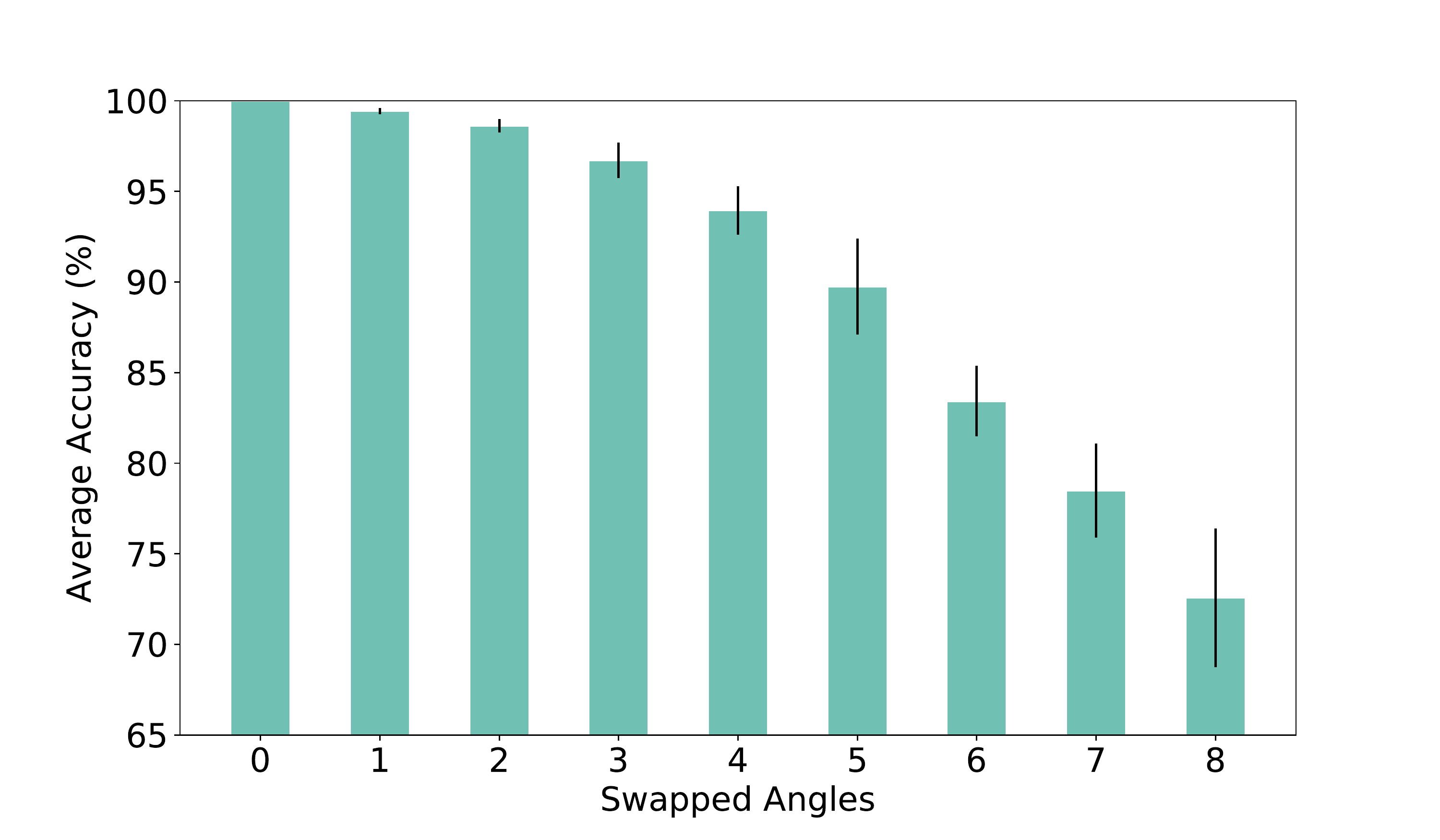}
 \caption{The performance of the orientation tracking approach when the orientation of the user is unknown. We feed data from specified number of angles to other encoders instead of their dedicated ones.}
  \label{fig:swapped_angle}
\end{figure}
To evaluate the performance of the orientation tracking approach when the orientation of the subject is unknown in the inference phase, we permute different number of angles. 
As shown in Fig.~\ref{fig:swapped_angle}, as we increase the number of angles for which the data is not fed to the dedicated encoder, the standard deviation of the average accuracy increases. 
Moreover, as expected, the accuracy decreases by increasing the number of swapped angles. 
As a result, it is crucial for orientation tracking approach to know the orientation of the user with respect to the NR sidelink operating radars.

\subsection{Angle Importance}
\label{sec:angle_importance}
\begin{figure*}
    \centering
    \includegraphics[width=.75\textwidth]{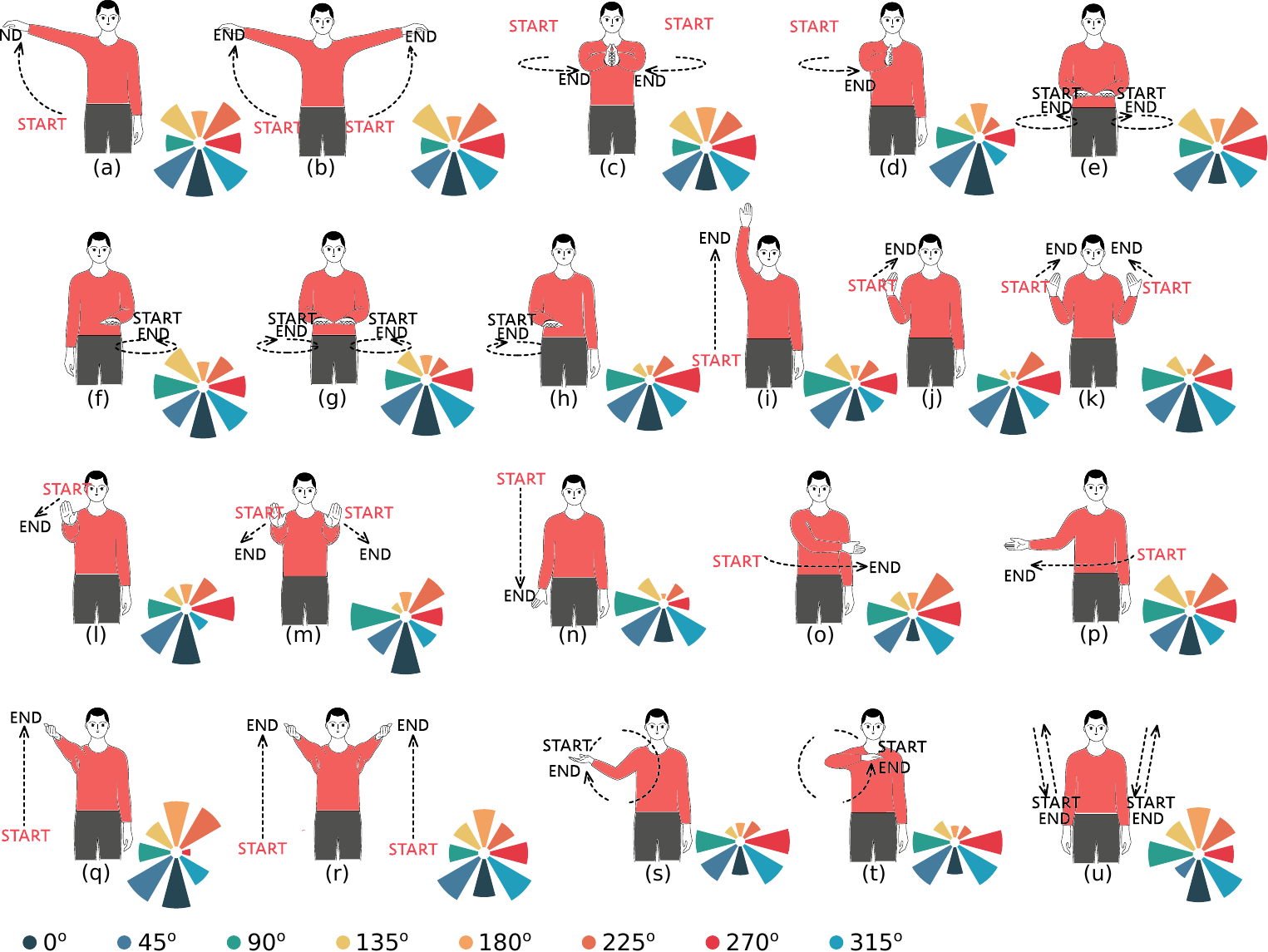}
    \caption{Gesture set used in the experiments:
        (a)~`lateral raise',
        (b)~`two-hand lateral-raise',
        (c)~`two-hand lateral-to-front',
        (d)~`lateral-to-front',
        (e)~`two-hand inward circles'
        (f)~`left-arm circle',
        (g)~`two-hand outward circles',
        (h)~`right-arm circle',
        (i)~`lift',
        (j)~`pull',
        (k)~`two-hand pull',
        (l)~`push',
        (m)~`two-hand push',
        (n)~`push-down',
        (o)~`swipe right',
        (p)~`swipe left',
        (q)~`throw',
        (r)~`two-hand throw',
        (s)~`circle counter-clockwise'.
        (t)~`circle clockwise',
        (u)~`arms swing'. In right side of each gesture, the importance of radars for the orientation tracking (the best) model is shown in a polar chart.
    }
    \label{fig:gesture_set}
\end{figure*}
Data coming from sidelink operating radars in different angles, differ even for the same gesture conducted due to the orientation change and the shadowing effect from a participant's body. 
In Fig.~\ref{fig:gesture_set}, the importance of each angle for each gesture in the orientation tracking approach is shown in a polar chart. 
For most gestures, the radars in front ($0^\circ$, $45^\circ$, and $315^\circ$) have the highest impact while the radar at the back ($180^\circ$) has the least importance. 
However, for few gestures like (q), (r), and (u) where the hands are visible from a back view, the impact of the radar at $180^\circ$ increases. 
Moreover, for the gestures that happen on one side of the body e.g. (d) and (n), the radars on the same side are  of higher importance compared to the radars on the other side.


\section{Conclusion}
We have proposed a mechanism for RF-convergence in cellular communication systems. 
In particular, we suggest to integrate RF-sensing with NR sidelink device-to-device communication, which is since release 12 part of the 3GPP cellular communication standards. 
We specifically investigated a common issue related to NR sidelink based RF-sensing, which is its angle and rotation dependence. 
In particular, we discussed transformations of mmWave point-cloud data which achieve rotational invariance, as well as distributed processing based on such rotational invariant inputs at distributed, angle and distance diverse devices.
Further, and to process the distributed data, we proposed a graph based encoder to capture spatio-temporal features of the data as well as four approaches for multi-angle learning. 
The approaches are compared on a newly recorded and openly available dataset comprising 15 subjects, performing 21 gestures which are recorded from 8~angles.
We were able to show that our data aggregation and processing toolchain outperforms the state-of-the-art point-cloud based gesture recognition approaches for angle-diverse gesture recordings.

%

%

\section*{Acknowledgment}
Part of the calculations presented above were performed using computer resources within the Aalto University School of Science “Science-IT” project.

This project has received funding from the European Union’s Horizon 2020 research and innovation programme under the Marie Skłodowska-Curie Grant agreement No. 813999.
We further appreciate partial funding through the ERANET-COFUND (H2020) CHIST-ERA III project RadioSense.

\ifCLASSOPTIONcaptionsoff
  \newpage
\fi



\bibliographystyle{IEEEtran}
\bibliography{references}
%



%

\begin{IEEEbiography}[{\includegraphics[width=1in,height=1.25in,clip,keepaspectratio]{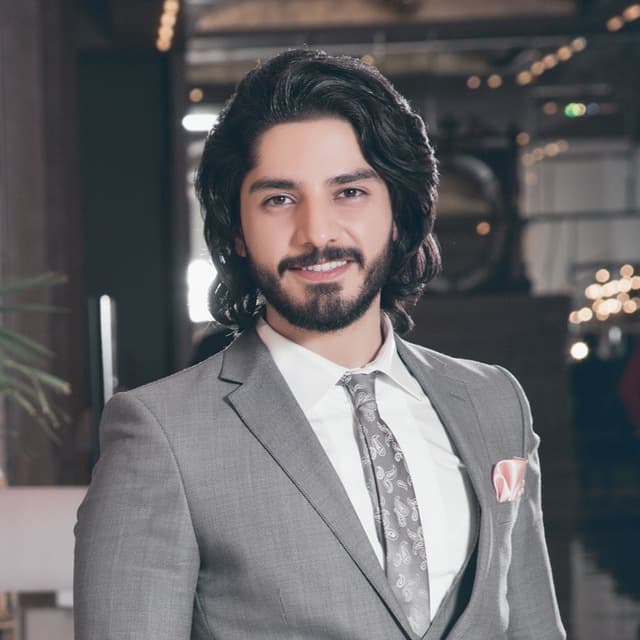}}]{Dariush Salami}
received his BSc and MSc degrees from Shahid Beheshti University and Amirkabir University of Technology in Software Engineering in 2016 and 2019, respectively. He is currently a Marie Skłodowska Curie fellow in ITN-WindMill project and a PhD researcher at the department of communications and networking at Aalto University. He is mainly focused on Machine Learning for Wireless Communications and Sensing especially in mmWave range.
\end{IEEEbiography}

\begin{IEEEbiography}[{\includegraphics[width=1in,height=1.25in,clip,keepaspectratio]{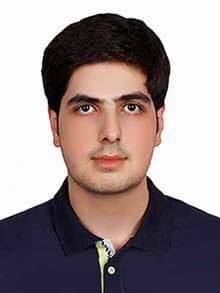}}]{Ramin Hasibi}
received his the BSc and MSc from Isfahan University of Technology and Amirkabir University of Technology in Information Technology Engineering in 2016 and 2019, respectively. He is currently a Ph.D. researcher at the department of informatics, University of Bergen where his main research focus is on Graph Representation Learning and Graph Neural Networks as well as their application in different domains.
\end{IEEEbiography}

\begin{IEEEbiography}[{\includegraphics[width=1in,height=1.25in,clip,keepaspectratio]{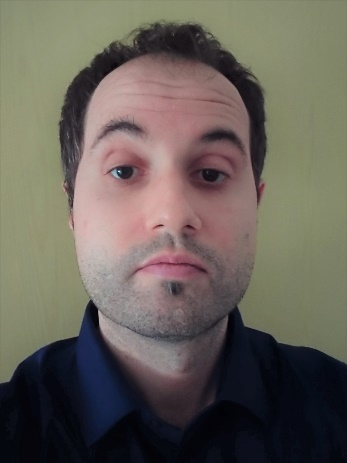}}]{Stefano Savazzi}
	received the M.Sc. degree and the Ph.D. degree (Hons.) in ICT from the Politecnico di Milano, Italy, in 2004 and 2008, respectively. In 2012, he joined the Institute of Electronics, Computer and Telecommunication Engineering (IEIIT), Consiglio Nazionale delle Ricerche (CNR), as a Researcher. He has coauthored over 100 scientific publications. His current research interests include distributed signal processing, learning and networking aspects for the Internet of Things, radio vision and localization. Dr. Savazzi won the Dimitris N. Chorafas Foundation Award in 2008. He is serving as Associate Editor for Frontiers in Communications and Networks and Topic Editor for Sensors (MDPI).
\end{IEEEbiography}

\begin{IEEEbiography}[{\includegraphics[width=1in,height=1.25in,clip,keepaspectratio]{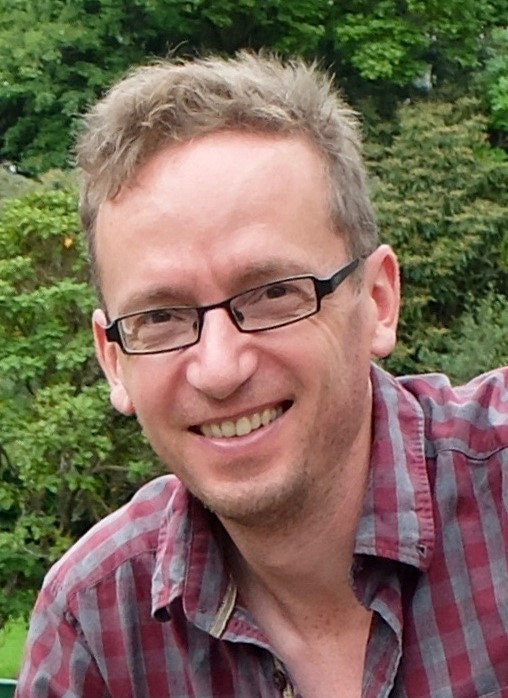}}]{Tom Michoel}
is Professor in bioinformatics at the Computational Biology Unit at the Department of Informatics at the University of Bergen since 2018, and was an independent group leader in computational biology at the University of Edinburgh (2012-2018) and the University of Freiburg (2010-2012). He obtained the MSc degree in Physics (1997) and PhD degree in Mathematical Physics (2001) from the KU Leuven, and was a postdoctoral researcher in mathematics (UC Davis, 2001-2002), theoretical physics (KU Leuven, 2002-2004), and bioinformatics and systems biology (Ghent University, 2004-2010). His research focus in the last five years has been on developing methods, algorithms, and software for causal inference and Bayesian network learning from high-dimensional omics data, supported by grants from the BBSRC (2015-2016), the NIH (2016-2019), the MRC (2017-2021), and the Norwegian Research Council (2021-2024).
\end{IEEEbiography}

\begin{IEEEbiography}[{\includegraphics[width=1in,height=1.25in,clip,keepaspectratio]{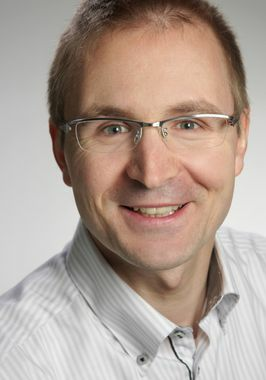}}]{Stephan Sigg}
 received his M.Sc. degree in computer science from TU Dortmund,
in 2004 and his Ph.D. degree from Kassel University, in 2008. Since 2015 he is
an assistant professor at Aalto University, Finland. He is a member of the
editorial board of the Proceedings of the ACM on Interactive, Mobile, Wearable
and Ubiquitous Technologies as well as of the Elsevier journal of Computer
Communications. He has served as a TPC member of renowned conferences
including IEEE PerCom, IEEE ICDCS, etc. His research interests include Ambient
Intelligence, in particular, Pervasive sensing, activity recognition, usable
security algorithms for mobile distributed systems.
\end{IEEEbiography}


\vfill


\end{document}